\documentclass[10pt,journal,cspaper,compsoc]{IEEEtran}
\usepackage{graphicx}
\usepackage{times}
\usepackage{amssymb,amsmath}
\usepackage{amsfonts}
\usepackage{fixmath}
\usepackage{bm}
\usepackage{amsmath}
\usepackage[bottom]{footmisc}
\usepackage{algpseudocode}
\usepackage{float}
\usepackage{multirow}
\usepackage[utf8]{inputenc}
 \usepackage{tikz}
 \usepackage{url}
 \usepackage[small]{caption}
 \usepackage{shortcuts}

\DeclareMathAlphabet{\mathpzc}{OT1}{pzc}{m}{it}

\DeclareMathOperator*{\pmS}{\,\pm\,}
\def\SDP{$\langle$SDP$\rangle$}
\def\SSP{$\langle$SSP$\rangle$}
\def\Ia{\rule[-0.0ex]{0ex}{2.5ex}}
\def\Ib{\rule[-0.5ex]{0ex}{3.0ex}} 
\begin{document}
\title{A Semidefinite Programming Based Search Strategy for Feature Selection with Mutual Information Measure}
\author{Tofigh~Naghibi \thanks{The authors are with the Computer Engineering and Networks Lab., ETH Zurich, Switzerland.

E-mail: \texttt{\{naghibi,hoffmann,pfister\}@tik.ee.ethz.ch}},
        Sarah~Hoffmann
        and~Beat~Pfister% <-this % stops a space
        }

\maketitle
\begin{abstract}
% --------------------------------
\label{Abstract}
Feature subset selection, as a special case of the general subset selection problem, has been the topic of a considerable number of studies due to the growing importance of data-mining applications. In the feature subset selection problem there are two main issues that need to be addressed: (i) Finding an appropriate measure function than can be fairly fast and robustly computed for high-dimensional data. (ii) A search strategy to optimize the measure over the subset space in a reasonable amount of time. In this article mutual information between features and class labels is considered to be the measure function. Two series expansions for mutual information are proposed, and it is shown that most heuristic criteria suggested in the literature are truncated
approximations of these expansions. It is well-known that searching the whole subset space is an
NP-hard problem. Here, instead of the conventional sequential search algorithms, we suggest a parallel search strategy based on semidefinite programming (SDP) that can search through the
subset space in polynomial time. By exploiting the similarities between the proposed algorithm and an instance of the maximum-cut problem in graph theory, the approximation ratio of this algorithm is derived and is compared with the approximation ratio of the backward elimination method. The experiments show that it can be misleading to judge the quality of a measure solely based on the classification accuracy, without taking the effect of the non-optimum search strategy into account. 
\end{abstract}
\section{Introduction}
%--------------------------------
\label{Introduction}
From a purely theoretical point of view, given the underlying conditional probability distribution of a dependent variable $C$ and a set of features $\mathbf{X}$, the Bayes decision rule can be applied to construct the optimum induction algorithm. However, in practice learning machines are not given access to this distribution, $Pr(C|\mathbf{X})$. Therefore, given a feature vector or variables $\mathbf{X}\in R^N$, the aim of most machine learning algorithms is to approximate this underlying distribution or estimate some of its characteristics. Unfortunately, in most practically relevant data mining applications, the dimensionality of the feature vector is quite high making it prohibitive to learn the underlying distribution. For instance, gene expression data or images may easily have more than tens of thousands of features. While, at least in theory, having more features should result in a more discriminative classifier, it is not the case in practice because of the computational burden and curse of 
dimensionality. 

High-dimensional data poses different challenges on induction and prediction algorithms. Essentially, the amount of data to sustain the spatial density of the underlying distribution increases exponentially with the dimensionality of the feature vector, or alternatively, the sparsity increases exponentially given a constant amount of data. Normally in real-world applications, a limited amount of data is available and obtaining a sufficiently good estimate of the underlying high-dimensional probability distribution is almost impossible unless for some special data structures or under some assumptions (independent features, etc). 

Thus, dimensionality reduction techniques, particularly feature extraction and feature selection methods, have to be employed to reconcile idealistic learning algorithms with real-world applications.  

In the context of feature selection, two main issues can be distinguished. The first one is to define an appropriate measure function to assign a score to a set of features. The second issue is to develop a search strategy that can find the optimal (in a sense of optimizing the value of the measure function) subset of features among all feasible subsets in a reasonable amount of time. 

Different approaches to address these two problems can roughly be categorized into three groups: Wrapper methods, embedded methods and filter methods.

Wrapper methods \cite{kohavi:96} use the performance of an induction algorithm (for instance a classifier) as the measure function. Given an inducer $\mathcal{I}$, wrapper approaches search through the space of all possible feature subsets and select the one that maximizes the induction accuracy. Most of the methods of this type require to check all the possible $2^N$ subsets of features and thus, may rapidly become prohibitive due to the so-called combinatorial explosion. Since the measure function is a machine learning (ML) algorithm, the selected feature subset is only optimal with respect to that particular algorithm, and may show poor generalization performance over other inducers. 

The second group of feature selection methods are called embedded methods \cite{neumann:04} and are based on some internal parameters of the ML algorithm. Embedded approaches rank features during the training process and thus simultaneously determine both the optimal features and the parameters of the ML algorithm. Since using (accessing) the internal parameters may not be applicable in all ML algorithms, this approach cannot be seen as a general solution to the  feature selection problem. In contrast to wrapper methods, embedded strategies do not require to run the exhaustive search over all subsets since they mostly evaluate each feature individually based on the score calculated from the internal parameters. However, similar to wrapper methods, embedded methods are dependent on the induction model and thus the selected subset is somehow tuned to a particular 
induction algorithm.

Filter methods, as the third group of selection algorithms, focus on filtering out irrelevant and redundant features in which irrelevancy is defined according to a predetermined measure function. Unlike the first two groups, filter methods do not incorporate the learning part and thus show better generalization power over a wider range of induction algorithms. They rely on finding an optimal feature subset through the optimization of a suitable measure function. Since the measure function is selected independently of the induction algorithm, this approach decouples the feature selection problem from the following ML algorithm. 

The first contribution of this work is to analyze the popular mutual information measure in the context of the feature selection problem. We will expand the mutual information function in two different series and show that most of the previously suggested information-theoretic criteria are the first or second order truncation-approximations of these expansions. The first expansion is based on generalization of mutual information and has already appeared in literature while the second one is new, to the best of our knowledge. The well-known minimal Redundancy Maximal Relevance (mRMR) score function can be immediately concluded from the second expansion. We will discuss the consistency and accuracy of these approximations and experimentally investigate the conditions in which these truncation-approximations may lead to high estimation errors.  

Alternatively, feature selection methods can be categorized based on the search strategies they employ. Popular search approaches can be divided into four categories: Exhaustive search, greedy search, projection and heuristic. A trivial approach is to exhaustively search in the subset space as it is done in wrapper methods. However, as the number of features increases, it can rapidly become infeasible. Hence, many popular search approaches use greedy hill climbing, as an approximation to this NP-hard combinatorial problem. Greedy algorithms iteratively evaluate a candidate subset of features, then modify the subset and evaluate if the new subset is an improvement over the old one. This can be done in a forward selection setup which starts with an empty set and adds one feature at a time or with a backward elimination process which starts with the full set of features and removes one feature at each step. The third group of the search algorithms are based on targeted projection pursuit which is a linear 
mapping 
algorithm to pursue an optimum projection of data onto a low dimensional manifold that scores highly with respect to a measure function \cite{friedman:74}. In heuristic methods, for instance genetic algorithms, the search is started with an initial subset of features which gradually evolves toward better solutions.

Recently, two convex quadratic programing based methods, QPFS in \cite{rod:10} and SOSS in \cite{naghibi:13} have been suggested to address the search problem. QPFS is a deterministic algorithm and utilizes the Nystr\"{o}m method to approximate large matrices for efficiency purposes. SOSS on the other hand, has a randomized rounding step which injects a degree of randomness into the algorithm in order to generate more diverse feature sets.     

Developing a new search strategy is another contribution of this paper. Here, we introduce a new class of search algorithms based on Semi-Definite Programming (SDP) relaxation. We reformulate the feature selection problem as a (0-1)-quadratic integer programming and will show that it can be relaxed to an SDP problem, which is convex and hence can be solved with efficient algorithms \cite {boyd:04}. Moreover, there is a discussion about the approximation ratio of the proposed algorithm in subsection 3.2. We show that it usually gives better solutions than greedy algorithms in the sense that its approximate solution is more probable to be closer to the optimal point of the criterion.
\vspace{-3mm}
\section{Mutual Information Pros and Cons}
%---------------------------------
\label{mutualinformation}
Let us consider an $N$ dimensional feature vector $\mathbf{X}=[X_1,X_2,...,X_N]$ and a dependent variable $C$ which can be either a class label in case of classification or a target variable in case of regression. The mutual information  function is defined as a distance from independence between  $\mathbf{X}$ and $C$ measured by the Kullback-Leibler divergence \cite{cover:91}. Basically, mutual information measures the amount of information shared between $\mathbf{X}$ and $C$ by measuring their dependency level. Denote the joint pdf of $\mathbf{X}$ and $C$ and its marginal distributions by $Pr(\mathbf{X},C)$, $Pr(\mathbf{X})$ and  $Pr(C)$, respectively. The mutual information between the feature vector and the class label can be defined as follows:
\begin{align} 
\label{eq_1} 
I(X_1,X_2,\dots,& X_N;C)\! = I(\mathbf{X};C) = \notag \\
& \! \int \!Pr(\mathbf{X},C)\! \log{\frac{Pr(\mathbf{X},C)}{Pr(\mathbf{X})Pr(C)}}\,\mathrm{d} \mathbf{X}\,\mathrm{d}C
\end{align}
It reaches its maximum value when the dependent variable is perfectly described by the feature set. In this case mutual information is equal to $H(C)$, where $H(C)$ is the Shannon entropy of $C$.

Mutual information can also be considered a measure of set intersection \cite{reza:61}. Namely, let $\mathbb{A}$ and $\mathbb{B}$ be event sets corresponding to random variables $A$ and $B$, respectively. It is not difficult to verify that a function $\mu$ defined as:
\begin{equation} 
\label{eq_2}
\mu({\mathbb{A} \cap \mathbb{B}}) = I(A;B)
\end{equation}
satisfies all three properties of a formal measure over sets \cite{yeung:91} \cite{bog:07}, i.e., non-negativity, assigning zero to empty set and countable additivity. However, as we see later, the generalization of the mutual information measure to more than two sets will no longer satisfy the \textit{non-negativity} property and thus can be seen as a signed measure which is the generalization of the concept of measure by allowing it to have negative values.   

There are at least three reasons for the popularity of the use of mutual information in feature selection algorithms. 

1. Most of the suggested non information-theoretic score functions are not formal set measures (for instance correlation function). Therefore, they cannot assign a score to a set of features but rather to individual features. However, mutual information as a formal set measure is able to evaluate all possible informative interactions and complex functional relations between features and as a result, fully extract the information contained in a set of features.

2. The relevance of the mutual information measure to misclassification error is supported by the existence of bounds relating the probability of misclassification of the Bayes classifier, $P_e$, to the mutual information. More specifically, Fano's weak lower bound \cite{fano:61} on $P_e$,
\begin{equation} 
\label{eq_3}
1+P_e\text{log}_2(n_y{-}1)\ge H(C)-I(\mathbf{X};C)
\end{equation}
where $n_y$ is the number of classes and the Hellman-Raviv \cite{hellman:70} upper bound,
\begin{equation} 
\label{eq_4}
P_e\le \frac{1}{2}(H(C)-I(\mathbf{X};C))
\end{equation}
on $P_e$, provide somewhat a performance guarantee. 

As it can be seen in \eref{eq_3} and \eref{eq_4}, maximizing the mutual information between $\mathbf{X}$ and $C$ decreases both upper and lower bounds on misclassification error and guarantees the goodness of the selected feature set. However, there is somewhat of a misunderstanding of this fact in the literature. It is sometimes wrongly claimed that maximizing the mutual information results in minimizing the $P_e$ of the optimal Bayes classifier. This is an unfounded claim since $P_e$ is not a monotonic function of the mutual information. Namely, it is possible that a feature vector $\mathbf{A}$ with less relevant information-content about the class label $C$ than a feature vector $\mathbf{B}$ yields a lower classification error rate than $\mathbf{B}$. The following example may clarify this point.

\textbf{Example 1}: Consider a binary classification problem with equal number of positive and negative training samples and two binary features $X_1$ and $X_2$. The goal is to select the optimum feature for the classification task. Suppose the first feature $X_1$ is positive if the outcome is positive. However, when the outcome is negative, $X_1$ can take both positive and negative values with the equal probability. Namely, $Pr(X_1{=}1|C{=}1) = 1$ and $Pr(X_1{=} -1 | C {=} -1) = 0.5$. In the same manner, the likelihood of $X_2$ is defined as $Pr(X_2 {=} 1 | C {=}1) = 0.9$ and $Pr(X_2 {=} -1 | C{=} -1) = 0.7$. Then, the Bayes classifier  with feature $X_1$ yields the classification error:
\begin{align} 
\label{ex_1}
P_{e1}= & Pr(C{=}{-}1)Pr(X_1{=}1|C{=}{-}1) \notag \\ 
        & +Pr(C{=}1)Pr(X{=}{-}1|C{=}1)=0.25 
\end{align}
Similarly, the Bayes classifier with $X_2$ yields $P_{e1}=0.2$ meaning that, $X_2$ is a better feature than $X_1$ in the sense of minimizing the probability of misclassification. However, unlike their error probabilities, $I(X_1;C) = 0.31$, is greater than 
$I(X_2;C) = 0.29$. That is, $X_1$ conveys more information about the class label in the sense of Shannon mutual information than $X_2$.

A more detailed discussion can be found in \cite{ben:12}. However, it is worthwhile to mention that although using mutual information may not necessarily result in the highest classification accuracy, it guarantees to reveal a salient feature subset by reducing the upper and lower bounds of $P_e$.

3- By adapting classification error as a criterion, most standard classification algorithms fail to correctly classify the instances from minority classes in imbalanced datasets. Common approaches to address this issue are to either assign higher misclassification costs to minority classes or replace the classification accuracy criterion with the area under the ROC curve which is a more relevant criterion when dealing with imbalanced datasets. Either way, the features should also be selected by an algorithm which is insensitive (robust) with respect to class distributions (otherwise the selected features may not be informative about minority classes, in the first place). Interestingly, by internally applying unequal class dependent costs, mutual information provides some robustness with respect to class distributions. Thus, even in an imbalanced case, a mutual information based feature selection algorithm is likely (though not guaranteed) to not overlook the features that represent the minority classes. In \
cite{bao:11}, the concept of the mutual information classifier is investigated. Specifically, the internal cost matrix of the mutual information classifier is derived to show that it applies unequal misclassification 
costs when dealing with imbalanced data and showed that the mutual information classifier is an optimal classifier in the sense of maximizing a weighted classification accuracy rate. The following example shows this robustness.

\textbf{Example 2}: Assume an imbalanced binary classification task where $Pr(C{=}1)=0.9$. As in Example 1, there are two binary features $X_1$ and $X_2$ and the goal is to select the optimum feature. Suppose $Pr(X_1{=}1|C{=}1) = 1$ and $Pr(X_1 {=} -1 | C {=} -1) = 0.5$. Unlike  the first feature, $X_2$ can much better classify the minority class $Pr(X_2 {=} {-}1 | C {=}{-}1) = 1$ and $Pr(X_2 {=}1 | C {=} 1) = 0.8$. It can be seen that the Bayes classifier with $X_1$ results in 100\% classification rate for the majority class while only 50\% correct classification for the minority. On the other hand, using $X_2$ leads to 100\%  correct classification for the minority class and 80\% for the majority. Based on the probability of error, $X_1$ should be preferred since its probability of error is  $P_{e1} = 0.05$ while  $P_{e2} = 0.18$. However, by using $X_1$ the classifier can not learn the rare event (50\% classification rate) and thus randomly classifies the minority class which is the class of interest in 
many applications. Interestingly, unlike the Bayesian error probabilities, mutual information prefers $X_2$ over $X_1$, since $I(X_2;C) = 0.20$ is greater than $I(X_1;C) = 0.18$. That is, mutual information is to some extent robust against imbalanced data.

Unfortunately, despite the theoretical appeal of the mutual information measure, given a limited amount of data, an accurate estimate of the mutual information would be impossible. Because to calculate mutual information, estimating the high-dimensional joint probability $Pr(\mathbf{X},C)$ is inevitable which is, in turn, known to be an NP hard problem \cite{karger:01}.

As mutual information is hard to evaluate, several alternatives have been suggested \cite{battiti:94}, \cite{peng:05}, \cite{kwak:02}. For instance, the Max-Relevance criterion approximates \eref{eq_1} with the sum of the mutual information values between individual features $X_i$ and $C$:
\begin{equation} 
\label{eq_5}
\text{Max-Relevance} = \sum_{i=1}^{N} I(X_i;C)
\end{equation}
Since it implicitly assumes that features are independent, it is likely that selected features are highly redundant. To overcome this problem, several heuristic corrective terms have been introduced to remove the redundant information and select mutually exclusive features. Here, it is shown that most of these heuristics are derived from the following expansions of mutual information with respect to $X_i$. 

\subsection{First Expansion: Multi-way Mutual Information Expansion}

The first expansion of mutual information that is used here, relies on the natural extension of mutual information to more than two random variables proposed by McGill \cite{mcgill:54} and Abramson \cite{abramson:63}. According to their proposal, the three-way mutual information between random variables $Y_i$ is defined by:
\begin{align} 
\label{eq_6}
I(Y_1;Y_2;Y_3) = & I(Y_1;Y_3)+I(Y_2;Y_3)-I(Y_1,Y_2;Y_3) \notag\\
               = &  I(Y_1;Y_2) - I(Y_1;Y_2|Y_3)
\end{align}
where ``,'' between variables denotes the joint variables. Note that, similar to two-way mutual information, it is symmetric with respect to $Y_i$ variables, i.e., $ I(Y_1;Y_2;Y_3) = I(Y_2;Y_3;Y_1)$. Generalizing over $N$ variables:
\begin{align} 
\label{eq_7}
I(Y_1;Y_2;\dots;Y_N) =\,  & I(Y_1;\dots;Y_{N-1}) \notag \\
                        &- I(Y_1;\dots;Y_{N-1}|Y_N)
\end{align}
Unlike 2-way mutual information, the generalized mutual information is not necessarily nonnegative and hence, can be interpreted as a signed measure of set intersection \cite{han:80}. Consider \eref{eq_6} and assume $Y_3$ is class label $C$, then positive $I(Y_1;Y_2;C)$ implies that $Y_1$ and $Y_2$ are redundant with respect to $C$ since $I(Y_1,Y_2;C) \le I(Y_1;C)+I(Y_2;C)$. However, the more interesting case is when $I(Y_1;Y_2;C)$ is negative, i.e., $I(Y_1,Y_2;C) \ge I(Y_1;C)+I(Y_2;C)$. This means, the information contained in the interactions of the variables is greater than the sum of the information of the individual variables \cite {gurban:09}.

\begin{figure}
\centering
\vspace{-0mm}
\includegraphics[scale=.45]{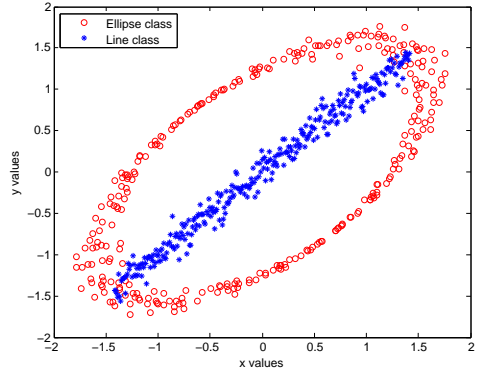}
\vspace{0mm}
\caption{Synergy between $x$ and $y$ features. While information of each individual feature about the class label (ellipse or line) is almost zero, their joint information can almost completely remove the class label ambiguity.}
\label{fig1}
\vspace{-6mm}
\end{figure}

An artificial example for this situation is the binary classification problem depicted in Figure \ref{fig1}, where the classification task is to discriminate between the 
ellipse class (class samples depicted by circles) and the line class (star samples) by using two features: values of $x$ axis and values of $y$ axis. As can be seen, since $I(x;C)\! \approx \! 0$ and $I(y;C)\! \approx\! 0$, there is no way to distinguish between these two classes by just using one of the features. However, it is obvious that employing both features results in almost perfect classification, i.e., $I(x,y;C)\! \approx\! H(C)$.
The mutual information in \eref{eq_1} can be expanded out in terms of generalized mutual information between the features and the class label as:
\begin{align} 
\label{eq_8}
I(\mathbf{X};C) =& \sum_{i_1=1}^{N} I(X_{i_1};C) -  \sum_{i_1=1}^{N-1} \sum_{i_2=i_1+1}^{N} I(X_{i_1};X_{i_2};C)  \notag  \\
   &  +\dots +   (-1)^{N-1} I(X_1;\dots;X_N;C) 
\end{align}
From the definition in \eref{eq_7} it is straightforward to infer this expansion. However, the more intuitive proof is to use the fact that mutual information is a measure of set intersection, i.e.,  $I(Y_1;Y_2;Y_3) = \mu(\mathbb{Y_1}\cap \mathbb{Y_2}\cap\mathbb{Y_3})$, where $\mathbb{Y_i}$ is the corresponding event set of the $Y_i$ variable. Now, expanding the $N$-variable measure function results in:
\begin{align} 
\label{eq_9}
I(\mathbf{X};C&) = \mu((\bigcup_{i=1}^{N} \mathbb{X}_i) \cap \mathbb{C}) =  \mu(\bigcup_{i=1}^{N} (\mathbb{X}_i \cap \mathbb{C})) \\
      &=  \sum_{i=1}^{N} \mu(\mathbb{X}_i \cap \mathbb{C}) -   \sum_{i_1=1}^{N-1} \sum_{i_2=i_1+1}^{N} \mu(\mathbb{X}_{i_1}\cap \mathbb{X}_{i_2} \cap \mathbb{C})  \notag  \\  
      & +\dots+ (-1)^{N-1}\mu(\mathbb{X}_1\cap \mathbb{X}_2 \dots \cap \mathbb{X}_N \cap \mathbb{C})   \notag
\end{align}
where the last equation follows directly from the addition law or sum rule in set theory. The proof is complete by recalling that all measure functions with the set intersection arguments in the last equation can be replaced by the mutual information functions according to the definition of mutual information in \eref{eq_2}.
\subsection {Second Expansion: Chain Rule of Information}

The second expansion for mutual information is based on the \textit{chain rule of information} \cite{cover:91}:
\begin{align} 
\label{eq_10}
I(\mathbf{X};C) = \sum_{i=1}^{N} I(X_i;C|X_{i-1},\dots,X_1)
\end{align}
The chain rule of information leaves the choice of ordering quite flexible. For example, the right side can be written in the order $(X_1,X_2,\dots,X_N)$ or $(X_N,X_{N-1},\dots,X_1)$. In general, it can be expanded over $N!$ different permutations of the feature set $\{X_1,\dots,X_N\}$. Taking the sum over all possible expansions yields,
\begin{align} 
\label{eq_10.5}
(N!)&I(\mathbf{X};C) =  (N{-}1)! \sum_{i=1}^{N} I(X_i;C) \\ 
                    &   + (N{-}2)! \sum_{i_1=1}^{N} \sum_{i_2{\in}\{1,...,N\}/i_1} I(X_{i_2};C|X_{i_1}) \notag \\
                    & + \cdots + (N{-}1)! \sum_{i=1}^{N} I(X_{i};C|\{X_1,\dots,X_N\}_{\backslash{X_{i}}}) \notag
\end{align}
Dividing both sides by $(N{-}1)!/2$, and using the following equation $I(X_{i_1};C|X_{i_2})= I(X_{i_1};C)-I(X_{i_1};X_{i_2};C)$ to replace $I(X_{i_1};C|X_{i_2})$ terms, our second expansion can be expressed as
\begin{align} 
\label{eq_11}
\frac{N}{2}& I( \mathbf{X};C)  = \sum_{i=1}^{N} I(X_{i};C)  \\
& -\frac{1}{N-1} \sum_{i_1=1}^{N-1} \sum_{i_2=i_1+1}^{N} I(X_{i_1};X_{i_2};C)  \notag \\
   &+ \dots + \frac{1}{2} \sum_{i=1}^{N} I(X_{i};C|\{X_1,\dots,X_N\}_{\backslash{X_{i}}}) \notag
 \end{align}
 Ignoring the unimportant multiplicative constant $N/2$ on the left side of equation \eref{eq_11}, the right side can be seen as a series expansion form of mutual information (up to a known constant factor). 
 
 \subsection{Truncation of the Expansions}
 In the both proposed expansions \eref{eq_8} and \eref{eq_11}, mutual information terms with more than two features represent higher-order interaction properties. Neglecting the higher order terms yields the so-called truncated approximation of the mutual information function. If we ignore the constant coefficient in \eref{eq_11}, the truncated forms of suggested expansions can be written as:
\begin{align} 
\label{eq_12}
D_1 = & \sum_{i=1}^{N} I(X_{i};C) -  \sum_{i=1}^{N-1} \sum_{j=i+1}^{N} I(X_{i};X_{j};C)\notag \\ 
D_2 = & \sum_{i=1}^{N} I(X_{i};C) -\frac{1}{N-1}  \sum_{i=1}^{N-1} \sum_{j=i+1}^{N} I(X_{i};X_{j};C)   
\end{align}
 where $D_1$ is the truncated approximation of \eref{eq_8} and $D_2$ is for \eref{eq_11}. Interestingly, despite the very similar structure of the expressions in \eref{eq_12}, they have intrinsically different behaviors. This difference seems to be rooted in different functional forms they employ to approximate the underlying high-order pdf with lower order distributions ( i.e., how they combine these lower order terms). For instance, the functional form that MIFS employs to approximate $Pr(\mathbf{x})$ is shown in \eref{kirk}. While $D_1$ is not necessarily a positive value, $D_2$ is guaranteed to be a positive approximation since all terms  in \eref{eq_10.5} are positive. However, $D_2$ may highly underestimate the mutual information values since it may violate the fact that \eref{eq_1} is always greater than or equal to $\max_{i} {I(X_i;C)}$.
 \subsubsection{JMI, mRMR \& MIFS Criteria}
 Several known criteria including Joint Mutual Information (JMI) \cite{meyer:06}, minimal Redundancy Maximal Relevance (mRMR) \cite{peng:05} and Mutual Information Feature Selection (MIFS) \cite{battiti:94} can immediately be derived from $D_1$ and $D_2$. 
 
 Using the identity: $I(X_i;X_j;C) = I(X_i;C)+I(X_j;C)-I(X_i,X_j;C)$ in $D_2$ reveals that $D_2$ is equivalent to JMI.
 \begin{align} 
\label{jmi}
\text{JMI} \!=D_2=  \sum_{i=1}^{N-1}\! \sum_{j=i+1}^{N} \!I(X_{i},X_{j};C)    
\end{align}

 Using $I(X_{i};X_{j};C) =I(X_{i};X_{j}) -I(X_{i};X_{j}|C)$ and ignoring the terms containing more than two variables, i.e., $I(X_{i};X_{j}|C)$, in the second approximation $D_2$, one may immediately recognize the popular score function
\begin{align} 
\label{eq_13}
\text{mRMR} \!= \!\sum_{i=1}^{N} I(X_{i};C)\! -\!\frac{1}{N-1}\!  \sum_{i=1}^{N-1}\! \sum_{j=i+1}^{N} \!I(X_{i};X_{j})    
\end{align}
introduced by Peng et al. in \cite{peng:05}. That is, mRMR is a truncated approximation of mutual information and not a heuristic approximation as suggested in \cite{brown:12}. 

The same line of reasoning as for mRMR can be applied to $D_1$ to achieve MIFS with $\beta$ equal to 1. 
\begin{align} 
\label{eq_13.5}
\text{MIFS} \!= \!\sum_{i=1}^{N} I(X_{i};C)\! -  \sum_{i=1}^{N-1}\! \sum_{j=i+1}^{N} \!I(X_{i};X_{j})    
\end{align} 

\textbf{Observation}:  A constant feature is a potential danger for the above measures. While adding an informative but correlated feature may reduce the score value (since $I(X_i;X_j|C)-I(X_i;X_j)$ can be negative), adding a non-informative constant feature $Z$ to a feature set does not reduce its score value since both $I(Z;C)$ and $I(Z;X_i;C)$ terms are zero, that is, constant features may be preferred over informative but correlated features. Therefore, it is essential to remove constant features by some preprocessing before using the above criteria for feature selection. 
\subsubsection{Implicitly Assumed Distribution}
 A natural question arising in this context with respect to the proposed truncated approximations is: Under what probabilistic assumptions do the proposed approximations become valid mutual information functions? That is, which structure should a joint pdf admit, to yield mutual information in the forms of $D_1$ or $D_2$?  
 
 For instance, if we assume features are mutually and class conditionally independent, i.e., $Pr(\mathbf{X}) = \prod_{i=1}^{N}Pr(X_i)$ and $Pr(\mathbf{X},C) = Pr(C)\prod_{i=1}^{N}Pr(X_i|C)$, then it is easy to verify that mutual information has the form of Max-Relevance introduced in \eref{eq_5}. These two assumptions, define the adapted \textit{independence-map} of $Pr(\mathbf{X},C)$ where the independence-map of a joint probability distribution is defined as follows.
 
 \textbf{Definition 1}: \textit{An independence-map (i-map) is a look up table or a set of rules that denote all the conditional and unconditional independence between random variables. Moreover, an i-map is consistent if it leads to a valid factorized probability distribution}.
 
 That is, given a consistent i-map, a high-order joint probability distribution is approximated with product of low-order pdfs and the obtained approximation is a valid pdf itself (e.g., $ \prod_{i=1}^{N}Pr(X_i)$ is an approximation of the high-order pdf $Pr(\mathbf{X})$ and it is also a valid probability distribution). 
 
 The question regarding the implicit consistent i-map that MIFS adopts has been investigated in \cite{balagani:10}. However, the assumption set (i-map) suggested in their work is inconsistent and leads to the incorrect conclusion that MIFS upper bounds the Bayesian classification error via the inequality \eref{eq_4}. As we show in the following theorem, unlike the Max-Relevance case, there is no i-map that can produce mutual information in the forms of mRMR of MIFS (ignoring the trivial solution that reduces mRMR or MIFS to Max-Relevance). 
 
 \textbf{Theorem 1.} \textit{Ignoring the trivial solution, i.e., the i-map indicating that random variables are mutually and class conditionally independent, there is no consistent i-map that can produce mutual information functions in the forms of mRMR \eref{eq_13} or MIFS \eref{eq_13.5} for arbitrary number of features}.
 
 \textbf{Proof}: The proof is by contradiction. Suppose there is a consistent i-map, where its corresponding joint pdf $\hat{P}r(\mathbf{X},C)$ (which is the approximation of $Pr(\mathbf{X},C)$) can generate mutual information in the forms of \eref{eq_13} or \eref{eq_13.5}. That is, if this i-map is adopted, by replacing $\hat{P}r(\mathbf{X},C)$ in \eref{eq_1} we get mRMR or MIFS. This implies that mRMR and MIFS are \textit{always} valid set measures for all datasets regardless of their true underlying joint probability distributions. Now, if we show (by any example) that they are not valid mutual information measures, i.e., they are not always positive and monotonic, then we have contradicted our assumption that $\hat{P}r(\mathbf{X},C)$ exists and is a valid pdf. It is not so difficult to construct an example in which mRMR or MIFS can get negative values. Consider the case where features are independent of class label, $I(X_i;C)=0$, while they have nonzero dependencies among themselves, $I(X_i;X_j)\neq0$. 
In this case, both mRMR and MIFS 
generate negative values which is not allowed by a valid set measure. This contradicts our assumption that they are generated by a valid distribution, so we are forced to conclude that there is no consistent i-map that results in mutual information in the mRMR or MIFS forms.$\blacksquare$
 
 The same line of reasoning can be used to show that $D_1$ and $D_2$ are also not valid measures. 
  
 However, despite the fact that no valid pdf can produce mutual information of those forms, it is still valid to ask for which low-order approximations of the underlying high-order pdfs, mutual information reduces to a truncated approximation form. That is, we do not restrict an approximation to be a valid distribution anymore. Any functional form of low-order pdfs may be seen as an approximation of the high-order pdfs and may give rise to MIFS or mRMR. In the next subsection we reveal these assumptions for the MIFS criterion. 
 \subsubsection{MIFS Derivation from Kirkwood Approximation}
 
 It is shown in \cite{killian:07} that truncation of the joint entropy $H(\mathbf{X})$ at the $r$th-order is equivalent to approximating the full-dimensional pdf $Pr(\mathbf{X})$ using joint pdfs with dimensionality of $r$ or smaller. This approximation is called $r$th order Kirkwood approximation. The truncation order that we choose, partially determines our belief about the structure of the function that we are going to estimate the exact $Pr(\mathbf{X})$ with.
 
 The 2nd order Kirkwood approximation of $Pr(\mathbf{X})$, can be denoted as follows \cite{killian:07}: 
  \begin{equation} 
 \label{kirk}
\hat{Pr}(\mathbf{X})= \frac{\prod_{i=1}^{N-1}\prod_{j=i+1}^{N}Pr(X_i,X_j)}{\big[\prod_{i=1}^{N}Pr(X_i)\big]^{N-2}}
 \end{equation} 
 Now, assume the following two assumptions hold:
 
 \textbf{Assumption 1}: Features are class conditionally independent, that is: $Pr(\mathbf{X}|C)=\prod_{i=1}^{N}Pr(X_i|C)$
 
 \textbf{Assumption 2}: $Pr(\mathbf{X})$ is well approximated by a 2nd order Kirkwood superposition approximation in \eref{kirk}.
 
 Then, writing the definition of mutual information and applying the above assumptions yields the MIFS criterion
   \begin{align} 
 \label{identity}
I(\mathbf{X};C) &= H(\mathbf{X})-H(\mathbf{X}|C) \\
		&  \stackrel{(a)}{\approx} \sum_{i=1}^{N} H(X_i) - \sum_{i=1}^{N-1}\sum_{j=i+1}^{N} I(X_i;X_j) -H(\mathbf{X}|C) \notag \\
		&  \stackrel{(b)}{=} \sum_{i=1}^{N} I(X_i;C) - \sum_{i=1}^{N-1}\sum_{j=i+1}^{N} I(X_i;X_j)  \notag
 \end{align} 
 In the above equation, (a) follows the second assumption by substituting the 2nd order Kirkwood approximation \eref{kirk} inside the logarithm of the entropy integral and (b) is an immediate consequence of the first assumption.
 
 The first assumption has already appeared in previous works \cite{brown:12} \cite{balagani:10}. However, the second assumption is novel and, to the best of our knowledge, the connection between the Kirkwood approximation and the MIFS criterion has not been explored before. 

 It is worth to mention that, in reality, both assumptions can be violated. Specifically, the Kirkwood approximation may not precisely reproduce dependencies we might observe in real-world datasets. Moreover, it is important to remember that the Kirkwood approximation is not a valid probability distribution.
 \vspace{-1mm}
 \subsection{$\mathbf{D}_2$ Approximation}
 From our experiments,  which we omit because of space constraints, $D_2$ tends to underestimate the mutual information while $D_1$ shows a large overestimation for independent features and a large underestimation (even becoming negative) in the presence of dependent features. In general, $D_2$ shows more robustness than $D_1$. The same results can be observed for mRMR which is derived from $D_2$ and MIFS derived from $D_1$. Previous work also arrived to the same results and reported that mRMR performs better and more robustly than MIFS especially when the feature set is large. Therefore, in the following sections we use $D_2$ as the truncated approximation. For simplicity, its subscript is dropped and it is rewritten as follows:   
\begin{align} 
\label{eq_14}
D(\{X_1,\dots,X_N\}) = &\sum_{i=1}^{N}  I(X_{i};C)  \\
- & \frac{1}{N-1} \sum_{i=1}^{N-1} \sum_{j=i+1}^{N} I(X_{i};X_{j};C)\notag
\end{align}
Note that although $D$ in \eref{eq_14} is not a formal set measure any more, it still can be seen as a score function for sets. However, it is noteworthy that unlike formal measures, the suggested approximations are no longer monotonic where the monotonicity merely means that a subset of features should not be better than any larger set that contains the very same subset. Therefore, as explained in \cite{narendra:77} the branch and bound based search strategies can not be applied to them. 

A very similar approach has been applied \cite{brown:09} (by using $D_1$ approximation) to derive several known criteria like MIFS \cite{battiti:94} and mRMR \cite{peng:05}. However, in \cite{brown:09} and most of other previous works, the set score function in \eref{eq_14} is immediately reduced to an individual-feature score function by fixing $N{-}1$ features in the feature set. This will let them to run a greedy selection search method over the feature set which essentially is a one-feature-at-a-time selection strategy. It is clearly a naive approximation of the optimal NP-hard search algorithm and may perform poorly under some conditions. In the following, we investigate a convex approximation of the binary objective function appearing in feature selection inspired by the Goemans-Williamson maximum cut approximation approach \cite{goemans:95}.

\section{Search Strategies}
%----------------------------
\label{search}
Given a measure function\footnote{By some abuse of terminology, we refer to any set function in this section as a measure, no matter whether they satisfy the formal measure properties.} $D$, the Subset Selection Problem (SSP) can be defined as follows:

\textbf{Definition 2}: Given $N$ features $X_i$ and a dependent variable $C$, select a subset of $P \! \ll \! N$ features that maximizes the measure function. Here it is assumed that the cardinality $P$ of the optimal feature subset is known.

In practice, the exact value of $P$ can be obtained by evaluating subsets for different values of cardinality $P$ with the final induction algorithm. Note that it is intrinsically different than wrapper methods. While in wrapper methods $2^N$ subsets have to be tested, here at most $N$ runs of the learning algorithm are needed to evaluate all possible values of $P$.  

A search strategy is an algorithm trying to find a feature subset in the feature subset space with $2^N$ members\footnote{Given a $P$, the size of the feature subset space reduces to  ${N\choose P}$.} that optimizes the measure function. The wide range of proposed search strategies in the literature can be divided into three categories: 1- Exponential complexity methods including exhaustive search \cite{kohavi:96}, branch and bound based algorithms \cite{narendra:77}. 2- Sequential selection strategies with two very popular members, forward selection and backward elimination methods. 3- Stochastic methods like simulated annealing and genetic algorithms \cite{vafai:93}, \cite{doak:92}.

Here, we introduce a fourth class of search strategies which is based on the convex relaxation of the 0-1 integer programming and explore its approximation ratio by establishing a link between SSP and an instance of the maximum-cut problem in graph theory. In the following, we briefly discuss the two popular sequential search methods and continue with the proposed solution: a close to optimal polynomial-time complexity search algorithm and its evaluation on different datasets.   
\vspace{-1mm}
\subsection{Convex Based Search}
%----------------------------------------------------
\label{sdp}
The forward selection (FS) algorithm selects a set $\mathbb{S}$ of size $P$ iteratively as follows:
\begin{enumerate}
\item Initialize $\mathbb{S}_0 = \emptyset$. 
\item In each iteration $i$, select the feature $X_m$ maximizing $D(\mathbb{S}_{i{-}1}\cup X_m)$, and set $\mathbb{S}_{i} = \mathbb{S}_{i{-}1} \cup {X_m}$. 
\item Output $\mathbb{S}_P$. 
\end{enumerate}
Similarly, backward elimination (BE) can be described as: 
\begin{enumerate}
\item Start with the full set of feature $\mathbb{S}_N$. 
\item Iteratively remove a variable $X_m$ maximizing $D(\mathbb{S}_i\backslash{X_m})$, and set $\mathbb{S}_{i-1} =\mathbb{S}_i\backslash{X_m}$, where removing $X$ from $\mathbb{S}$ is denoted by $\mathbb{S}\backslash{X}$.
\item Output $\mathbb{S}_P$.
\end{enumerate}
An experimentally comparative evaluation of several variants of these two algorithms has been conducted in \cite{aha:94}. From an information theoretical standpoint, the main disadvantage of the forward selection method is that it only can evaluate the utility of a single feature in the limited context of the previously selected features. The artificial binary classifier in Figure \ref{fig1} may illustrate this issue. Since the information content of each feature ($x$ and $y$) is almost zero, it is highly probable that the forward selection method fails to select them in the presence of some other more informative features. 

Contrary to forward selection, backward elimination can evaluate the contribution of a given feature in the context of all other features. Perhaps this is why it has been frequently reported to show superior performance than forward selection. However, its overemphasis on feature interactions is a double-edged sword and may lead to a sub-optimal solution. 

\textbf{Example 3}: Imagine a four dimensional feature selection problem where $X_1$ and $X_2$ are class conditionally and mutually independent of $X_3$ and $X_4$, i.e., $Pr(X_1,X_2,X_3,X_4)=Pr(X_1,X_2)Pr(X_3,X_4)$ and $Pr(X_1,X_2,X_3,X_4|C)=Pr(X_1,X_2|C)Pr(X_3,X_4|C)$. Consider $I(X_1;C)$ and $I(X_2;C)$ are equal to zero, while their interaction is informative. That is, $I(X_1,X_2;C) = 0.4$. Moreover, assume $I(X_3;C) = 0.2$, $I(X_4;C)=0.25$ and $I(X_3,X_4;C) = 0.45$. The goal is to select only two features out of four. Here, backward elimination will select $\{X_1,X_2\}$ rather than the optimal subset $\{X_3,X_4\}$ because, removing any of $X_1$ or $X_2$ features will result in $0.4$ reduction of the mutual information value $I(X_1,\dots,X_4;C)$, while eliminating $X_3$ or $X_4$ deducts at most $0.25$. One may draw the conclusion that backward elimination tends to sacrifice the individually-informative features in favor of the merely cooperatively-informative features. As a remedy, several hybrid 
forward-backward sequential search methods have been proposed. However, they all fail in one way or another and more importantly cannot guarantee the goodness of the solution.

Alternatively, a sequential search method can be seen as an approximation of the combinatorial subset selection problem. To propose a new approximation method, the underlying combinatorial problem has to be studied. To this end, we may formulate the SSP defined in the beginning of this section as:
 \begin{align}
\label{eq_15}
 &\max_{\mathbf{x}} \, {\mathbf{x}^T\mathbf{Q}\mathbf{x}}  \notag \\
  & \sum_{i=1}^{N} x_i = P \\ 
 & x_i \in \{0,1\} \text {   for   } i=1,\dots,N \notag
 \end{align}
 where $\mathbf{Q}$ is a symmetric mutual information matrix constructed from the mutual information terms in \eref{eq_14}:
 \begin{equation} 
\label{eq_16}
  \mathbf{Q} = \begin{pmatrix}
   I(X_1;C) &  \cdots & -\frac{\lambda}{2}I(X_1;X_N;C)\\
   -\frac{\lambda}{2}I(X_1;X_2;C) & \cdots &  -\frac{\lambda}{2}I(X_2;X_N;C)\\
  \vdots   & \ddots & \vdots  \\
  -\frac{\lambda}{2}I(X_1;X_N;C) & \cdots & I(X_N;C)\\
\end{pmatrix}
 \end{equation}
where $\lambda =\frac{1}{P-1}$ and $\mathbf{x} = [x_1,\dots,x_N]$ is a binary vector where the variables $x_i$ are set-membership binary variables indicating the presence of the corresponding features $X_i$ in the feature subset.
It is straightforward to verify that for any binary vector $\mathbf{x}$, the objective function in \eref{eq_15} is equal to the score function $D(\mathbb{X}_{nz})$ where $\mathbb{X}_{nz} =\{X_i | x_i =1;i=1,\dots,N \}$. 
Note that, for mRMR $I(X_i;X_j;C)$ terms have to be replaced with $I(X_i;X_j)$.

The (0,1)-quadratic programming problem \eref{eq_15} has attracted a great deal of theoretical study because of its importance in combinatorial problems \cite[and references therein]{poljak:95}. This problem can simply be transformed to a \text{(-1,1)-quadratic} programming problem,
\begin{align} 
\label{eq_17}
 &\max_{\mathbf{y}} {\frac{1}{4}\mathbf{y}^T\mathbf{Q}\mathbf{y} + \frac{1}{2}\mathbf{y}^T\mathbf{Q}\mathbf{e} + c}  \notag \\
 & \sum_{i=1}^{N} y_i = 2P-N \\ 
 & y_i \in \{-1,1\} \text {   for   } i=1,\dots,N \notag
 \end{align}
via $\mathbf{y} = 2\mathbf{x}-\mathbf{e}$ transformation, where $\mathbf{e}$ is an all ones vector. Additionally $c$ in the above formulation is a constant equal to $\frac{1}{4}\mathbf{e}^T\mathbf{Qe}$ and it can be ignored because of its independence of $\mathbf{y}$. In order to homogenize the objective function in \eref{eq_17}, define an $(N{+}1)\!\times\!(N{+}1)$ matrix $\mathbf{Q}^u$ by adding a 0-th row and column to $\mathbf{Q}$ so that:
\begin{align} 
\label{eq_18}
\mathbf{Q}^u = \begin{pmatrix}
  0 & \mathbf{e}^T\mathbf{Q}\\
  \mathbf{Q}^T\mathbf{e} & \mathbf{Q}
\end{pmatrix} 
 \end{align}
 Ignoring the constant factor $\frac{1}{4}$ in \eref{eq_17}, the equivalent homogeneous form of \eref{eq_15} can be written as:
\begin{align} 
\label{eq_19}
 & S_{\text{SSP}} = \max_{\mathbf{y}} {\mathbf{y}^T\mathbf{Q}^u\mathbf{y}}  \notag \\
  \text{\SSP}\quad \quad  & \sum_{i=1}^{N} y_iy_0 = 2P-N\\ 
 & y_i \in \{-1,1\} \text {   for   } i=0,\dots,N \notag
\end{align}
Note that $\mathbf{y}$ is now an $N+1$ dimensional vector with the first element $y_0 = \pm 1$ as a reference variable. Given the solution of the problem above, i.e., $\mathbf{y}$, the optimal feature subset is obtained by $\mathbb{X}_{op} =\{X_i|y_i=y_0\}$. 

The optimization problem in \eref{eq_19} can be seen as an instance of the maximum-cut problem \cite{goemans:95} with an additional cardinality constraint, also known as the k-heaviest subgraph or maximum partitioning graph problem. The two main approaches to solve this combinatorial problem are either to use the linear programming relaxation by linearizing the product of two binary variables \cite{frieze:83}, or the semidefinite programming (SDP) relaxation suggested in \cite {goemans:95}. The SDP relaxation has been proved to have
exceptionally high performance and achieves the approximation ratio of 0.878 for the original maximum-cut problem. The SDP relaxation of \eref{eq_19} is:
\begin{align} 
\label{eq_20}
 &S_{\text{SDP}} = \max_{\mathbf{Y}} {\text{tr}\{\mathbf{Q}^u\mathbf{Y}\}}  \notag \\
 & \sum_{i,j=1}^{N} Y_{ij} = (2P-N)^2  \notag \\
\text{\SDP}\quad \quad & \sum_{i=1}^{N} Y_{i0} = (2P-N)  \\
 & \text{diag}(\mathbf{Y}) =\mathbf{e}  \notag \\
 & \mathbf{Y}\succeq 0   \notag
\end{align}
where $\mathbf{Y}$ is an unknown $(N+1)\times(N+1)$ positive semidefinite matrix and
$\text{tr}\{\mathbf{Y}\}$ denotes its trace. Obviously, any feasible solution $\mathbf{y}$ for
\SSP is also feasible for its SDP relaxation by $\mathbf{Y} =
\mathbf{y}\mathbf{y}^T$. Furthermore, it is not difficult to see that any rank one solution,
$\text{rank}(\mathbf{Y})=1$, of \SDP is a solution of \SSP. 

 The \SDP problem can be solved within an additive error $\gamma$ of the optimum by for example interior point methods \cite{boyd:04} whose computational complexity are polynomial in the size of the input and $log(\frac{1}{\gamma})$. However, since its solution is not necessarily a rank one matrix, we need some more steps to obtain a feasible solution for \SSP. The following three steps summarize the approximation algorithm for \SSP which in the following will be referred to as convex based relaxation approximation (COBRA) algorithm.

\vspace{2mm}
\noindent\textbf{COBRA Algorithm}:
\vspace{1.5mm}
\begin{enumerate}
\item SDP: Solve \SDP and obtain $\mathbf{Y}_{sdp}$. Repeat the following steps many times and output the best solution.

\item  Randomized rounding: Using the multivariate normal distribution with a zero mean and a covariance matrix $\mathbf{R} = \mathbf{Y}_{sdp}$ to sample $\mathbf{u}$ from distribution $\mathcal{N}(0,\mathbf{R})$ and construct $\hat{\mathbf{x}} = \text{sign}(\mathbf{u})$. Select $\mathbb{X} = \{X_i | \hat{x}_i = \hat{x}_0\}$.

\item Size adjusting: By using the greedy forward or backward algorithm, resize the cardinality of $\mathbb{X}$ to $P$. 
\end{enumerate}

The randomized rounding step is a standard procedure to produce a binary solution from the real-valued solution of \SDP and is widely used for designing and analyzing approximation algorithms \cite{ragh:88}. The third step is to construct a feasible solution that satisfies the cardinality constraint. Generally, it can be skipped since in feature selection problems the exact satisfaction of the cardinality constraint is not required.

We use the SDP-NAL solver \cite{zhao:10} with the Yalmip interface \cite{lof:04} to implement this algorithm in Matlab. SDP-NAL uses the Newton-CG augmented Lagrangian method to efficiently solve SDP problems. It can solve large scale problems ($N$ up to a few thousand) in an hour on a PC with an Intel Core i7 CPU. Even more efficient algorithms for low-rank SDP have been suggested claiming that they can solve problems with the size up to $N{=}30000$ in a reasonable amount of time \cite{grippo:12}. Here we only use the SDP-NAL for our experiments.     
\vspace{-3mm}
\subsection{Approximation Analysis}
\begin{center}
 \begin{table*}
 \addtolength{\tabcolsep}{5mm}
 \centering
   \begin{tabular}{l|| c c c c c c c} 
\hline\Ia
\bf{Values of $P$}     &  $N/2$     &  $N/3$ & $N/4$ & $N/6$ &$N/8$& $N/10$ &$N/20$
\\
\hline 
\hline
\ \bf{BE}            & 0.4         & 0.25   & 0.16    & 0.10    & 0.071    & 0.055    & 0.026 \Ia    \\

\ \bf{COBRA}          & 0.48       & 0.33   & 0.24    &0.13     & 0.082    &  0.056   & 0.015     \\ 
\hline
\end{tabular}
\caption{Approximation ratios of BE and COBRA for different $N/P$ values.}
\vspace{-2mm}
\label{app-ratio}
\end{table*}  
\end{center}
\vspace{-3mm}
In order to gain more insight into the quality of a measure function, it is essential to be able to directly examine it. However, since estimating the exact mutual information value in real data is not feasible, it is not possible to directly evaluate the measure function. Its quality can only be indirectly examined through the final classification performance (or other measurable criteria). However, the quality of a measure function is not the only contributor to the classification rate. Since SSP is an NP-hard problem, the search strategy can only find a local optimal solution. That is, besides the quality of a measure function, the inaccuracy of the search strategy also contributes to the final classification error. Thus, in order to draw a conclusion concerning the quality of a measure function, it is essential to have an insight about the accuracy of the search strategy in use. In this section, we compare the accuracy of the proposed method with the traditional backward elimination approach.

A standard approach to investigate the accuracy of an optimization algorithm is by analyzing how close it gets to the optimal solution. Unfortunately, feature selection is an NP-hard problem and thus achieving the optimal solution to use as reference is only feasible for small-sized problems. In such cases, one wants a provable solution's quality and certain properties about the algorithm, such as its approximation ratio. Given a maximization problem, an algorithm is called $\rho$-approximation algorithm if the approximate solution is at least $\rho$ times the optimal value. That is, in our case $\rho S_{SSP}\le S_{COBRA}$, where $S_{COBRA}=D(\mathbb{X}_{COBRA})$. The factor $\rho$ is usually referred to as the approximation ratio in the literature.  

The approximation ratios of BE and COBRA can be found by linking the SSP to the k-heaviest subgraph problem (k-HSP) in graph theory. k-HSP is an instance of the max-cut problem with a cardinality constraint on the selected subset, that is, to determine a subset $S$ of k vertices such that the weight of the subgraph induced by $S$ is maximized \cite{sriva:98}. From the definition of k-HSP, it is clear that SSP with the criterion \eref{eq_14} is equivalent to the $P$-heaviest subgraph problem since it selects the heaviest subset of features with the cardinality $P$, where heaviness of a set is the score assigned to it by $D$.

An SDP based algorithm for k-HSP has been suggested in \cite{sriva:98} and its approximation ratio has been analyzed. Their results are directly applicable to COBRA since both algorithms use the same randomization method (step 2 of COBRA) and the randomization is the main ingredient of their approximation analysis. The approximation ratio of BE for k-HSP has been investigated in \cite{asahiro:00}. It is a deterministic analysis and their results are also valid for our case, i.e., using BE for maximizing $D$. 

The approximation ratios of both algorithms for different values of $P$, as a function of $N$ (total number of features), have been listed in Table \ref{app-ratio} (values are calculated from the formulas in \cite{asahiro:00}). As can be seen, as $P$ becomes smaller, the approximation ratio approaches zero yielding the trivial lower bound 0 on the approximate solution. However, for larger values of $P$, the approximation ratio is nontrivial since it is bounded away from zero. For all cases shown in the table except the last one, COBRA gives better guarantee bound than BE. Thus, we may conclude that COBRA is more likely to achieve better approximate solution than BE.

In the following section, we will focus on comparing our search algorithm with sequential search methods in conjunction with different measure functions and over different classifiers and datasets.

\section{Experiments}

The evaluation of a feature selection algorithm is an intrinsically difficult task since there is no direct way to evaluate the goodness of a \textit{selection process} in general. Thus, usually a selection algorithm is scored based on the performance of its output, i.e., the selected feature subset, in some specific classification (regression) system. This kind of evaluation can be referred to as the goal-dependent evaluation. However, this method obviously cannot evaluate the generalization power of the selection process on different induction algorithms. To evaluate the generalization strength of a feature selection algorithm, we need a goal-independent evaluation. Thus, for evaluation of the feature selection algorithms, we propose to compare the algorithms over different datasets with multiple classifiers. This method leads to a more classifier-independent evaluation process. 

Some properties of the eight datasets used in the experiments are listed in Table \ref{datasets}. All datasets are available on the UCI machine learning archive \cite{frank:10}, except the NCI data which can be found in the website of Peng et al. \cite{peng:05}. These datasets have been widely used in previous feature selection studies \cite{peng:05}, \cite{ciarelli:10}. The goodness of each feature set is evaluated with five classifiers including Support Vector Machine (SVM), Random Forest (RF), Classification and Regression Tree (CART), Neural Network (NN) and Linear Discriminant Analysis (LDA). To derive the classification accuracies, 10-fold cross-validation is performed except for the NCI, DBW and LNG datasets where leave-one-out cross-validation is used.
  
As explained before, filter-based methods consist of two components: A measure function and a search strategy. The measure functions we use for our experiments are mRMR and JMI defined in \eref{eq_13} and \eref{jmi}, respectively. To unambiguously refer to an algorithm, it is denoted by measure function + search method used in that algorithm, eg., mRMR+FS. 
\vspace{-20mm}
\begin{center}
\begin{table*}
\addtolength{\tabcolsep}{1mm}
\centering
\begin{tabular}{l ||c  c  c  c  c  c  c  c } 
\hline \Ia
 \bf{Dataset Name}         & Arrhythmia & NCI             & DBWorld e-mails    & CNAE-9  & Internet Adv. & Madelon & Lung Cancer & Dexter          \\ 
  \bf{Mnemonic}        & ARR        & NCI                  & DBW        & CNA     & IAD           & MAD     &LNG          &DEX                             \\ \hline\hline
  \# \bf{Features}    & 278        & 9703                 &  4702      & 856     & 1558          & 500     &56           &20000                            \\   
  \# \bf{Samples}     & 370        & 60                   & 64         & 1080    & 3279          & 2000      & 32          &300                         \\   
  \# \bf{Classes}     & 2          & 9                    & 2          & 9       & 2             & 2       &3             &2                                \\  
  \hline
\end{tabular}
\caption{Datasets descriptions}
\label{datasets}
\end{table*}
\end{center}

\begin{table}
\def\nlsp{\\[1.6mm]}
\framebox{
\parbox{80mm}{
{\normalsize 
Set $P$: $\mathbb{P}\!=\!\{P_1,\dots,P_L\}$.
 \begin{algorithmic}
\ForAll {$P$ in $\mathbb{P}$}
  
     \State Run the COBRA algorithm and output the solution $\mathbb{X}$.
     \State Derive the classifier error-rate by applying K-fold 
     \State cross-validation and save it in $CL(P)$. 
   
\EndFor 
\end{algorithmic}
 Output: $ P_{opt} = \underset{P}{\operatorname{argmin}}\, CL(P)$
 }}
 }
 \vspace{-1mm}
 \caption{ Estimating $P$ by searching over an admissible set that minimizes the classification error-rate.}
 \label{algo}
\end{table} 

\begin{center}
  \begin{table*}[!ht]
  \addtolength{\tabcolsep}{3.5mm}
   \hfill{}
    \begin{tabular}{l||  c@{\hspace{3pt}}c c@{\hspace{3pt}}c c@{\hspace{3pt}}c c@{\hspace{3pt}}c c@{\hspace{3pt}}c c  } 
      \hline  \Ia 
   \textbf{Classifiers}                    &  \multicolumn{2}{|c|}{\bf{SVM}}                      &  \multicolumn{2}{|c|}{\bf{LDA}}                &  \multicolumn{2}{|c|}{\bf{CART}}            &\multicolumn{2}{|c|}{\bf{RF}}                     & \multicolumn{2}{|c|}{\bf{NN}}        & \bf{Average}            \\   \hline  \hline   
                                                                &   \multicolumn{11}{c}{\bf{NCI Dataset}}  	\Ia		                                             \\ 
   \bf{mRMR+COBRA}                     & (54) & 81.7               & (95) & 78.3         & (20) & 45.0          & (71)& 88.3           &  (60) & 75.0 & \textbf{73.67}                          \\  
   \bf{mRMR+FS}                        &  (32) & 78.3  	           &  (11) & 68.3        &  (2) & 45.0        &   (12) & 83.3        &  (99) & 70.0    & 69.00                \\ 
   \bf{mRMR+BE}                        &  (26) & 76.6              &  (11) & 68.3        &  (2) & 45.0        &   (13) & 85.0        &  (31) & 71.7    & 69.33                                \\  
    
   \bf{JMI+COBRA}                      &  (72)& 85.0              &  (70)& 75.0        &  (28)& 45.0       &   (45)& 90.0             &  (93)& 75.0      & 74.00                          \\  
   \bf{JMI+FS}                         &  (27)& 75.0              &  (17)& 68.3       &  (82)& 45.0       &   (17)& 86.6             &  (78)&  70.0     & 69.00                   \\  
   \bf{JMI+BE}                         &  (23)& 76.6              &  (20)&  76.6       &  (7)& 33.3      &   (19)& 86.6             &  (89)& 76.6     & 70.00                                \\  \hline\hline
   
                                                                  &   \multicolumn{11}{c}{ \bf{DBW Dataset}}  	\Ia		                                             \\ 
   \bf{mRMR+COBRA}                    &  (38)&   96.9             &  (152)&   92.2        &  (38)&   86.0         &    (33)&   92.2       &    (33)&  98.4  & \textbf{93.12}                    \\ 
   \bf{mRMR+FS}                       &  (31)&   93.7             &  (4)&   89.0        &  (4)&   86.0          &   (7)&    90.6       &    (9)&  92.2  & 90.31         \\ 
   \bf{mRMR+BE}                       &  (110) &   93.7             &  (6)     &   89.0        &  (4)   &   82.8          &   (29)   &   92.2       &    (9)   &  92.2  & 90.00          
   \\           
   \bf{JMI+COBRA}                     & (35)  & 93.7                  &  (14) & 89.0            &  (8)  & 82.8          &  (24)  & 92.2      &  (108)  & 93.7  & 90.31                         \\  
   \bf{JMI+FS}                        & (23) &93.7                    &  (6)  & 89.0            &  (5)  & 82.8          &  (34)  & 92.2      &  (96)  & 92.2   & 90.00                \\  
   \bf{JMI+BE}                        & (24) & 93.7                   &  (6) &  89.0            &  (5)  & 82.8          &  (23)  & 92.2      &  (149)  & 92.2  & 90.00                               \\  \hline\hline
    
                                                                     &   \multicolumn{11}{c}{ \bf{CNA Dataset}} 	\Ia		                                             \\ 
   \bf{mRMR+COBRA}                    &   (200)  & 94.0 &  (183)  &  92.7      &  (63)  & 75.0         &  (183) & 90.8       & (187)  & 92.0     & 88.91     \\  
   \bf{mRMR+FS}                       &   (149)  & 90.6 &  (142)  &  90.4      &  (7)   & 70.2         &  (138) & 87.7       & (78)   & 85.5     & 84.88        \\ 
   \bf{mRMR+BE}                       &   (199)  & 94.0 &  (165)  &  92.5      &  (47)  & 75.0         &  (176) & 90.8       & (84)   & 92.2     & 88.90    \\  
   
   \bf{JMI+COBRA}                     &  (140)  & 92.6  &  (146)  & 92.2  &  (47)  & 75.0    &  (148)  & 90.4  &  (148)  & 91.4  & 88.30                        \\  
   \bf{JMI+FS}                        &  (150)  & 92.7  &  (142)  & 92.1  &  (48)  & 75.3    &  (148)  & 90.7  &  (145)  & 91.3  & 88.40               \\ 
   \bf{JMI+BE}                        &  (150)  & 92.7  &  (142)  & 92.1  &  (48)  & 75.0    &  (144)  & 90.4  &  (134)  & 91.2  & 88.30                            \\  \hline\hline
   
                                                                       &   \multicolumn{11}{c}{ \bf{IAD Dataset}} \Ia		                                             \\
   \bf{mRMR+COBRA}                    & (165) & 96.5            &  (140) & 96.1      &  (28) & 96.4           &  (160) & 97.2      &   (68) & 97.1   	  & 96.64         \\ 
   \bf{mRMR+FS}                       & (109) & 96.2            &  (127) & 95.8      &  (127) & 96.7          &  (25)  & 97.0      &   (52) & 97.2	  & 96.58               \\ 
   \bf{mRMR+BE}                       & (22)  & 96.3            &  (163) & 95.9       &  (121) & 96.1          &  (109) & 97.2      &  (148) & 97.4  	  & 96.58             \\  
   
   \bf{JMI+COBRA}                     & (112)  & 96.3  &  (4)  & 96.3  &  (9)  & 96.3  &  (57)  & 97.3  &  (140)  & 100& 97.24                        \\ 
   \bf{JMI+FS}                        & (9)  & 96.2    &  (4)  & 96.2 &  (52)  & 96.4  &   (7)  & 96.8  &  (7)  & 97.8  & 96.68             \\  
   \bf{JMI+BE}                        & (4)  & 96.6    &  (17)  & 95.8  &  (79) & 96.3  &  (13) & 96.5  &  (10) & 97.2  & 96.48                            \\  \hline\hline
   
                                                                     &   \multicolumn{11}{c}{ \bf{MAD Dataset}}  	\Ia		                                             \\ 
   \bf{mRMR+COBRA}                    &   (12)  & 83.2  &  (13)  & 60.4  &  (26)  & 80.5  &  (12)  & 88.0 &  (11)  & 62.2    & \textbf{74.81}        \\  
   \bf{mRMR+FS}                       &   (32)  & 55.3  &  (5)   & 55.5  &  (12)  & 58.2  &  (49)  & 57.3  &  (5)    & 52.7  & \textbf{55.82}               \\  
   \bf{mRMR+BE}                       &   (14)  & 55.3  &  (11)  & 54.8  &  (31)  & 57.3  &  (26)  & 56.4  &  (115)  & 48.6  & 54.50             \\    
   
   \bf{JMI+COBRA}                     &  (13)  & 82.5  &  (12)  & 60.7  &  (40)  & 80.7  &  (13)  & 87.6  &  (4)   & 61.1  & 74.54                         \\  
   \bf{JMI+FS}                        &  (13)  & 82.5  &  (12)  & 60.7  &  (58)  & 80.5  &  (13)  & 87.9  &  (19)  & 59.2  & 74.20               \\  
   \bf{JMI+BE}                        &  (13)  & 82.5  &  (12)  & 60.7  &  (58)  & 80.5  &  (13)  & 87.3  &  (20)  & 60.1  & 74.25                               \\  \hline\hline
   
								    &   \multicolumn{11}{c}{ \bf{LNG Dataset}} 	\Ia		                                             \\  
   \bf{mRMR+COBRA}                    &  (23) & 75.0  &  (28) & 96.9   &  (13) & 71.8  &  (28) & 68.7  &  (27) & 71.8  & \textbf{76.87}         \\  
   \bf{mRMR+FS}                       &  (7)  & 81.2  &  (5)  & 68.7   &  (5)  & 71.8  &  (5)  & 75.0  &  (6)  & 71.8  & 73.75                 \\  
   \bf{mRMR+BE}                       &  (7)  & 81.2  &  (4)  & 68.7   &  (4)  & 71.8  &  (4)  & 75.0  &  (4)  & 75.0  & 74.37            \\ 
   
   \bf{JMI+COBRA}                     &  (7)  & 78.1       &   (6) & 71.8      &  (5)  & 71.8     &   (5)  & 75.0       &  (5) &  68.7 & 73.12                        \\  
   \bf{JMI+FS}                        &  (7)  & 78.1        &  (4) & 71.8       &  (4)  & 71.8     &  (8)  & 78.1       &  (5) &  68.7 & 73.75              \\ 
   \bf{JMI+BE}                        &  (7)  & 78.1        &  (6) & 71.8       &  (5)  & 71.8     &  (6)  & 78.1       &  (6) &  71.8 & 74.37                              \\  \hline\hline
   
								  &   \multicolumn{11}{c}{ \bf{ARR Dataset}} 	\Ia		                                             \\ 
   \bf{mRMR+COBRA}                    &   (45)  & 81.9  &  (48)  & 76.3  &  (30)  & 75.4  &  (43)  & 82.2  &  (57) & 72.9 & 77.75          \\
   \bf{mRMR+FS}                       &   (34)  & 81.3  &  (43)  & 76.1  &  (7)   & 78.3  &  (34)  & 81.3  &  (5)  & 75.7  & 78.56                \\
   \bf{mRMR+BE}                       &   (36)  & 81.6  &  (43)  & 76.3  &  (22)  & 78.0  &  (25)  & 82.9  &  (8)  & 76.1  & 79.02               \\ 
   
   \bf{JMI+COBRA}                     &   (26)  & 80.6  &  (51)  & 74.7  &  (15)  & 78.3  &  (51)  & 81.5  &  (13)  & 71.9       & 77.41                          \\  
   \bf{JMI+FS}                        &   (47)  & 74.3  &  (38)  & 73.5  &  (26)  & 76.9  &  (37)  & 79.2  &  (54)  & 70.0       & 74.80                \\ 
   \bf{JMI+BE}                        &   (47)  & 74.3  &  (38)  & 73.5  &  (26)  & 76.9  &  (25)  & 80.0  &  (29)  & 68.6       & 74.66                               \\  \hline\hline
   
   								    &   \multicolumn{11}{c}{ \bf{DEX Dataset}}  \Ia			                                             \\ 
   \bf{mRMR+COBRA}                    &    (3)  & 92.0    &  (131) & 86.3   &  (24)  & 80.7  &  (3)  & 93.0   &(3)   & 81.3            & 86.66         \\  
   \bf{mRMR+FS}                       &    (3)  & 90.3  &  (56)  & 87.0   &  (94)  & 80.3    &  (3)  & 92.0   &(3)   & 80.0            & 86.00               \\  
   \bf{mRMR+BE}                       &    (3)  & 90.0  &  (131)  & 87.3  &  (18)  & 80.3    &  (3)  & 91.6   &(99)  & 78.6            & 85.53              \\  
   
   \bf{JMI+COBRA}                     &  (88)   & 91.6  &  (13)   & 83.0  &  (12)  & 80.3    &  (3)    & 94.0  &  (3)   & 81.0          &   86.00                      \\ 
   \bf{JMI+FS}                        &  (149)  & 91.0  &  (129)  & 87.6  &  (95)  & 80.3    &  (119)  & 92.3  &  (94)  & 80.6          &   86.40              \\  
   \bf{JMI+BE}                        &  (149)  & 90.0  &  (128)  & 87.3  &  (22)  & 81.0    &  (146)  & 92.0  &  (138) &  78.0         &   85.60                             \\  \hline
   
  \end{tabular}
      \hfill{}
       \caption{Comparison of COBRA with the greedy search methods over different datasets. For each classifier and combination of search method and measure function, the values in parentheses is the number of selected features and the second value is the classification accuracy. The last column reports the average of the classification accuracies for each algorithm.}
       \label{ssp-be}
  \end{table*} 
 \end{center}
\vspace{0mm}

  A simple algorithm listed in Table \ref{algo} is employed to search for the optimal value of the subset cardinality $P$, where $P$ ranges over a set $\mathbb{P}$ of admissible values. In the worst case, $\mathbb{P} =\{1,\dots,N\}$. 

  Table \ref{ssp-be} shows the results obtained for the 8 datasets and 5 classifiers. Friedman test with the corresponding Wilcoxon-Nemenyi post-hoc analysis was used to compare the different algorithms. However, looking at the classification rates even before running the Friedman tests on them reveals a few interesting points which are marked in bold font.
  
  First, on the small size datasets (NCI, DBW and LNG), mRMR+COBRA consistently shows higher performance than other algorithms. The reason lies in the fact that the \textit{similarity ratio} of the feature sets selected by COBRA is lower than BE or FS feature sets. The \textit{similarity ratio} $S_i$ is defined as the number of features in the intersection of $i$th and $i{+}1$th feature sets divided by the cardinality of the $i$th feature set. From its definition it is clear that for BE and FS this ratio is always equal to 1. However, because of the randomization step this ratio may widely vary for COBRA. That is, COBRA generates 
quite diverse feature sets. Some of these feature sets have relatively low scores as compared with BE or FS sets. However, since for small datasets the estimated mutual information terms are highly inaccurate, features that rank low with our noisy measure function may in fact be better for classification. The average of the similarity ratios of 50 subsequent feature sets ($\frac{1}{50}\sum_{i=5}^{55} S_i$) have been reported for 4 datasets in Table \ref{s-ratio}. As seen, for NCI the averaged similarity ratio is significantly smaller than 1 while for CNA which is a relatively larger dataset, it is almost constant and equal to 1.
  
\begin{table}
     \addtolength{\tabcolsep}{2mm}
   \hfill{}
    \begin{tabular}{l|| c c c c} 
      \hline \Ia  
   \textbf{Datasets}                     &  \bf{NCI}            &     \bf{DBW}  &\bf{IAD}              &  \bf{CNA}                          \\ \hline  \hline  \Ia 
   \bf{S-ratio}                          &   0.7717             &     0.8929    & 0.9266               &   0.9976 \\ \hline 
     \end{tabular}
      \hfill{}
       \caption{The average (over 50 similarity ratios) similarity ratio for 4 datasets.}
       \label{s-ratio}
\end{table} 

  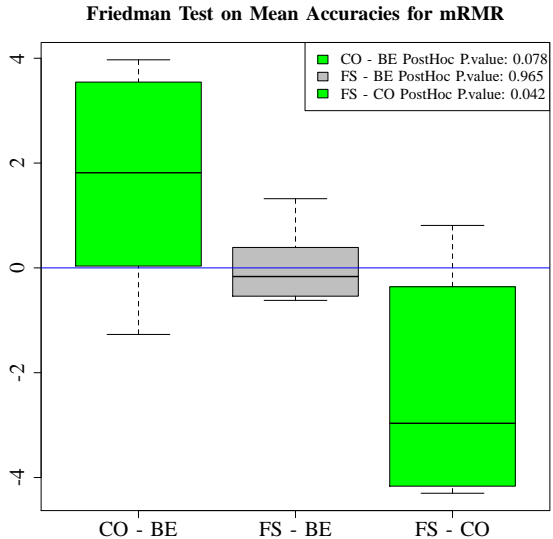
\begin{figure}
\resizebox{8cm}{!}{
   % Created by tikzDevice version 0.6.2-92-0ad2792 on 2013-12-19 21:24:33
% !TEX encoding = UTF-8 Unicode
\begin{tikzpicture}[x=1pt,y=1pt]
\definecolor[named]{fillColor}{rgb}{1.00,1.00,1.00}
\path[use as bounding box,fill=fillColor,fill opacity=0.00] (0,0) rectangle (505.89,505.89);
\begin{scope}
\path[clip] ( 49.20, 61.20) rectangle (480.69,456.69);
\definecolor[named]{fillColor}{rgb}{0.00,1.00,0.00}

\path[fill=fillColor] ( 78.50,267.80) --
	(185.04,267.80) --
	(185.04,423.22) --
	( 78.50,423.22) --
	cycle;
\definecolor[named]{drawColor}{rgb}{0.00,0.00,0.00}

\path[draw=drawColor,line width= 1.2pt,line join=round] ( 78.50,346.62) -- (185.04,346.62);

\path[draw=drawColor,line width= 0.4pt,dash pattern=on 4pt off 4pt ,line join=round,line cap=round] (131.77,210.02) -- (131.77,267.80);

\path[draw=drawColor,line width= 0.4pt,dash pattern=on 4pt off 4pt ,line join=round,line cap=round] (131.77,442.04) -- (131.77,423.22);

\path[draw=drawColor,line width= 0.4pt,line join=round,line cap=round] (105.13,210.02) -- (158.40,210.02);

\path[draw=drawColor,line width= 0.4pt,line join=round,line cap=round] (105.13,442.04) -- (158.40,442.04);

\path[draw=drawColor,line width= 0.4pt,line join=round,line cap=round] ( 78.50,267.80) --
	(185.04,267.80) --
	(185.04,423.22) --
	( 78.50,423.22) --
	( 78.50,267.80);
\definecolor[named]{fillColor}{rgb}{0.75,0.75,0.75}

\path[fill=fillColor] (211.67,242.34) --
	(318.22,242.34) --
	(318.22,283.52) --
	(211.67,283.52) --
	cycle;

\path[draw=drawColor,line width= 1.2pt,line join=round] (211.67,258.95) -- (318.22,258.95);

\path[draw=drawColor,line width= 0.4pt,dash pattern=on 4pt off 4pt ,line join=round,line cap=round] (264.94,238.80) -- (264.94,242.34);

\path[draw=drawColor,line width= 0.4pt,dash pattern=on 4pt off 4pt ,line join=round,line cap=round] (264.94,324.70) -- (264.94,283.52);

\path[draw=drawColor,line width= 0.4pt,line join=round,line cap=round] (238.31,238.80) -- (291.58,238.80);

\path[draw=drawColor,line width= 0.4pt,line join=round,line cap=round] (238.31,324.70) -- (291.58,324.70);

\path[draw=drawColor,line width= 0.4pt,line join=round,line cap=round] (211.67,242.34) --
	(318.22,242.34) --
	(318.22,283.52) --
	(211.67,283.52) --
	(211.67,242.34);
\definecolor[named]{fillColor}{rgb}{0.00,1.00,0.00}

\path[fill=fillColor] (344.85, 81.83) --
	(451.39, 81.83) --
	(451.39,250.31) --
	(344.85,250.31) --
	cycle;

\path[draw=drawColor,line width= 1.2pt,line join=round] (344.85,134.96) -- (451.39,134.96);

\path[draw=drawColor,line width= 0.4pt,dash pattern=on 4pt off 4pt ,line join=round,line cap=round] (398.12, 75.85) -- (398.12, 81.83);

\path[draw=drawColor,line width= 0.4pt,dash pattern=on 4pt off 4pt ,line join=round,line cap=round] (398.12,302.12) -- (398.12,250.31);

\path[draw=drawColor,line width= 0.4pt,line join=round,line cap=round] (371.49, 75.85) -- (424.76, 75.85);

\path[draw=drawColor,line width= 0.4pt,line join=round,line cap=round] (371.49,302.12) -- (424.76,302.12);

\path[draw=drawColor,line width= 0.4pt,line join=round,line cap=round] (344.85, 81.83) --
	(451.39, 81.83) --
	(451.39,250.31) --
	(344.85,250.31) --
	(344.85, 81.83);
\end{scope}
\begin{scope}
\path[clip] (  0.00,  0.00) rectangle (505.89,505.89);
\definecolor[named]{drawColor}{rgb}{0.00,0.00,0.00}

\path[draw=drawColor,line width= 0.4pt,line join=round,line cap=round] (131.77, 61.20) -- (398.12, 61.20);

\path[draw=drawColor,line width= 0.4pt,line join=round,line cap=round] (131.77, 61.20) -- (131.77, 55.20);

\path[draw=drawColor,line width= 0.4pt,line join=round,line cap=round] (264.94, 61.20) -- (264.94, 55.20);

\path[draw=drawColor,line width= 0.4pt,line join=round,line cap=round] (398.12, 61.20) -- (398.12, 55.20);

\node[text=drawColor,anchor=base,inner sep=0pt, outer sep=0pt, scale=  1.80] at (131.77, 39.60) {CO - BE};

\node[text=drawColor,anchor=base,inner sep=0pt, outer sep=0pt, scale=  1.80] at (264.94, 39.60) {FS - BE};

\node[text=drawColor,anchor=base,inner sep=0pt, outer sep=0pt, scale=  1.80] at (398.12, 39.60) {FS - CO};

\path[draw=drawColor,line width= 0.4pt,line join=round,line cap=round] ( 49.20, 89.13) -- ( 49.20,443.37);

\path[draw=drawColor,line width= 0.4pt,line join=round,line cap=round] ( 49.20, 89.13) -- ( 43.20, 89.13);

\path[draw=drawColor,line width= 0.4pt,line join=round,line cap=round] ( 49.20,177.69) -- ( 43.20,177.69);

\path[draw=drawColor,line width= 0.4pt,line join=round,line cap=round] ( 49.20,266.25) -- ( 43.20,266.25);

\path[draw=drawColor,line width= 0.4pt,line join=round,line cap=round] ( 49.20,354.81) -- ( 43.20,354.81);

\path[draw=drawColor,line width= 0.4pt,line join=round,line cap=round] ( 49.20,443.37) -- ( 43.20,443.37);

\node[text=drawColor,rotate= 90.00,anchor=base,inner sep=0pt, outer sep=0pt, scale=  1.80] at ( 34.80, 89.13) {-4};

\node[text=drawColor,rotate= 90.00,anchor=base,inner sep=0pt, outer sep=0pt, scale=  1.80] at ( 34.80,177.69) {-2};

\node[text=drawColor,rotate= 90.00,anchor=base,inner sep=0pt, outer sep=0pt, scale=  1.80] at ( 34.80,266.25) {0};

\node[text=drawColor,rotate= 90.00,anchor=base,inner sep=0pt, outer sep=0pt, scale=  1.80] at ( 34.80,354.81) {2};

\node[text=drawColor,rotate= 90.00,anchor=base,inner sep=0pt, outer sep=0pt, scale=  1.80] at ( 34.80,443.37) {4};
\end{scope}
\begin{scope}
\path[clip] (  0.00,  0.00) rectangle (505.89,505.89);
\definecolor[named]{drawColor}{rgb}{0.00,0.00,0.00}

\node[text=drawColor,anchor=base,inner sep=0pt, outer sep=0pt, scale=  1.70] at (264.94,475.42) {\bfseries Friedman Test on Mean Accuracies for mRMR};
\end{scope}
\begin{scope}
\path[clip] (  0.00,  0.00) rectangle (505.89,505.89);
\definecolor[named]{drawColor}{rgb}{0.00,0.00,0.00}

\path[draw=drawColor,line width= 0.4pt,line join=round,line cap=round] ( 49.20, 61.20) --
	(480.69, 61.20) --
	(480.69,456.69) --
	( 49.20,456.69) --
	( 49.20, 61.20);
\end{scope}
\begin{scope}
\path[clip] ( 49.20, 61.20) rectangle (480.69,456.69);
\definecolor[named]{drawColor}{rgb}{0.00,0.00,0.00}

\path[draw=drawColor,line width= 0.4pt,line join=round,line cap=round] (273.03,456.69) rectangle (480.69,399.09);
\definecolor[named]{fillColor}{rgb}{0.00,1.00,0.00}

\path[draw=drawColor,line width= 0.4pt,line join=round,line cap=round,fill=fillColor] (283.83,445.89) rectangle (292.47,438.69);
\definecolor[named]{fillColor}{rgb}{0.75,0.75,0.75}

\path[draw=drawColor,line width= 0.4pt,line join=round,line cap=round,fill=fillColor] (283.83,431.49) rectangle (292.47,424.29);
\definecolor[named]{fillColor}{rgb}{0.00,1.00,0.00}

\path[draw=drawColor,line width= 0.4pt,line join=round,line cap=round,fill=fillColor] (283.83,417.09) rectangle (292.47,409.89);

\node[text=drawColor,anchor=base west,inner sep=0pt, outer sep=0pt, scale=  1.30] at (303.27,438.16) {CO - BE  PostHoc P.value: 0.078};

\node[text=drawColor,anchor=base west,inner sep=0pt, outer sep=0pt, scale=  1.30] at (303.27,423.76) {FS - BE  PostHoc P.value: 0.965};

\node[text=drawColor,anchor=base west,inner sep=0pt, outer sep=0pt, scale=  1.30] at (303.27,409.36) {FS - CO  PostHoc P.value: 0.042};
\definecolor[named]{drawColor}{rgb}{0.00,0.00,1.00}

\path[draw=drawColor,line width= 0.4pt,line join=round,line cap=round] ( 49.20,266.25) -- (480.69,266.25);
\end{scope}
\end{tikzpicture}

   }
\caption{Comparing the search strategies for mRMR measure with the Friedman test and its corresponding post-hoc analysis. The y-axis is the classification accuracy difference and x-axis indicates the names of the compared algorithms.}
\label{mrmr-mean}
\end{figure}

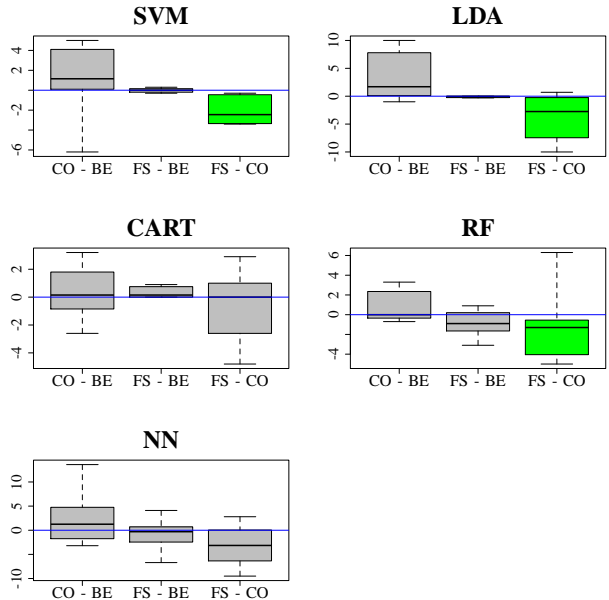
\begin{figure}
\resizebox{85mm}{!}{
   % Created by tikzDevice version 0.6.2-92-0ad2792 on 2013-12-19 19:09:46
% !TEX encoding = UTF-8 Unicode
\begin{tikzpicture}[x=1pt,y=1pt]
\definecolor[named]{fillColor}{rgb}{1.00,1.00,1.00}
\path[use as bounding box,fill=fillColor,fill opacity=0.00] (0,0) rectangle (505.89,505.89);
\begin{scope}
\path[clip] ( 32.47,377.65) rectangle (236.31,473.42);
\definecolor[named]{fillColor}{rgb}{0.75,0.75,0.75}

\path[fill=fillColor] ( 46.31,431.08) --
	( 96.64,431.08) --
	( 96.64,462.75) --
	( 46.31,462.75) --
	cycle;
\definecolor[named]{drawColor}{rgb}{0.00,0.00,0.00}

\path[draw=drawColor,line width= 1.2pt,line join=round] ( 46.31,439.39) -- ( 96.64,439.39);

\path[draw=drawColor,line width= 0.4pt,dash pattern=on 4pt off 4pt ,line join=round,line cap=round] ( 71.48,381.20) -- ( 71.48,431.08);

\path[draw=drawColor,line width= 0.4pt,dash pattern=on 4pt off 4pt ,line join=round,line cap=round] ( 71.48,469.87) -- ( 71.48,462.75);

\path[draw=drawColor,line width= 0.4pt,line join=round,line cap=round] ( 58.90,381.20) -- ( 84.06,381.20);

\path[draw=drawColor,line width= 0.4pt,line join=round,line cap=round] ( 58.90,469.87) -- ( 84.06,469.87);

\path[draw=drawColor,line width= 0.4pt,line join=round,line cap=round] ( 46.31,431.08) --
	( 96.64,431.08) --
	( 96.64,462.75) --
	( 46.31,462.75) --
	( 46.31,431.08);

\path[fill=fillColor] (109.23,428.70) --
	(159.56,428.70) --
	(159.56,431.47) --
	(109.23,431.47) --
	cycle;

\path[draw=drawColor,line width= 1.2pt,line join=round] (109.23,430.29) -- (159.56,430.29);

\path[draw=drawColor,line width= 0.4pt,dash pattern=on 4pt off 4pt ,line join=round,line cap=round] (134.39,427.91) -- (134.39,428.70);

\path[draw=drawColor,line width= 0.4pt,dash pattern=on 4pt off 4pt ,line join=round,line cap=round] (134.39,432.66) -- (134.39,431.47);

\path[draw=drawColor,line width= 0.4pt,line join=round,line cap=round] (121.81,427.91) -- (146.98,427.91);

\path[draw=drawColor,line width= 0.4pt,line join=round,line cap=round] (121.81,432.66) -- (146.98,432.66);

\path[draw=drawColor,line width= 0.4pt,line join=round,line cap=round] (109.23,428.70) --
	(159.56,428.70) --
	(159.56,431.47) --
	(109.23,431.47) --
	(109.23,428.70);
\definecolor[named]{fillColor}{rgb}{0.00,1.00,0.00}

\path[fill=fillColor] (172.14,403.76) --
	(222.47,403.76) --
	(222.47,426.72) --
	(172.14,426.72) --
	cycle;

\path[draw=drawColor,line width= 1.2pt,line join=round] (172.14,410.89) -- (222.47,410.89);

\path[draw=drawColor,line width= 0.4pt,dash pattern=on 4pt off 4pt ,line join=round,line cap=round] (197.31,403.37) -- (197.31,403.76);

\path[draw=drawColor,line width= 0.4pt,dash pattern=on 4pt off 4pt ,line join=round,line cap=round] (197.31,427.91) -- (197.31,426.72);

\path[draw=drawColor,line width= 0.4pt,line join=round,line cap=round] (184.72,403.37) -- (209.89,403.37);

\path[draw=drawColor,line width= 0.4pt,line join=round,line cap=round] (184.72,427.91) -- (209.89,427.91);

\path[draw=drawColor,line width= 0.4pt,line join=round,line cap=round] (172.14,403.76) --
	(222.47,403.76) --
	(222.47,426.72) --
	(172.14,426.72) --
	(172.14,403.76);
\end{scope}
\begin{scope}
\path[clip] (  0.00,  0.00) rectangle (505.89,505.89);
\definecolor[named]{drawColor}{rgb}{0.00,0.00,0.00}

\path[draw=drawColor,line width= 0.4pt,line join=round,line cap=round] ( 71.48,377.65) -- (197.31,377.65);

\path[draw=drawColor,line width= 0.4pt,line join=round,line cap=round] ( 71.48,377.65) -- ( 71.48,373.69);

\path[draw=drawColor,line width= 0.4pt,line join=round,line cap=round] (134.39,377.65) -- (134.39,373.69);

\path[draw=drawColor,line width= 0.4pt,line join=round,line cap=round] (197.31,377.65) -- (197.31,373.69);

\node[text=drawColor,anchor=base,inner sep=0pt, outer sep=0pt, scale=  1.3] at ( 71.48,363.40) {CO - BE};

\node[text=drawColor,anchor=base,inner sep=0pt, outer sep=0pt, scale=  1.3] at (134.39,363.40) {FS - BE};

\node[text=drawColor,anchor=base,inner sep=0pt, outer sep=0pt, scale=  1.3] at (197.31,363.40) {FS - CO};

\path[draw=drawColor,line width= 0.4pt,line join=round,line cap=round] ( 32.47,382.78) -- ( 32.47,461.95);

\path[draw=drawColor,line width= 0.4pt,line join=round,line cap=round] ( 32.47,382.78) -- ( 28.51,382.78);

\path[draw=drawColor,line width= 0.4pt,line join=round,line cap=round] ( 32.47,398.62) -- ( 28.51,398.62);

\path[draw=drawColor,line width= 0.4pt,line join=round,line cap=round] ( 32.47,414.45) -- ( 28.51,414.45);

\path[draw=drawColor,line width= 0.4pt,line join=round,line cap=round] ( 32.47,430.29) -- ( 28.51,430.29);

\path[draw=drawColor,line width= 0.4pt,line join=round,line cap=round] ( 32.47,446.12) -- ( 28.51,446.12);

\path[draw=drawColor,line width= 0.4pt,line join=round,line cap=round] ( 32.47,461.95) -- ( 28.51,461.95);

\node[text=drawColor,rotate= 90.00,anchor=base,inner sep=0pt, outer sep=0pt, scale=  1.2] at ( 22.97,382.78) {-6};

\node[text=drawColor,rotate= 90.00,anchor=base,inner sep=0pt, outer sep=0pt, scale=  1.2] at ( 22.97,414.45) {-2};

\node[text=drawColor,rotate= 90.00,anchor=base,inner sep=0pt, outer sep=0pt, scale=  1.2] at ( 22.97,446.12) {2};

\node[text=drawColor,rotate= 90.00,anchor=base,inner sep=0pt, outer sep=0pt, scale=  1.2] at ( 22.97,461.95) {4};
\end{scope}
\begin{scope}
\path[clip] (  0.00,337.26) rectangle (252.94,505.89);
\definecolor[named]{drawColor}{rgb}{0.00,0.00,0.00}

\node[text=drawColor,anchor=base,inner sep=0pt, outer sep=0pt, scale=  1.98] at (134.39,482.82) {\bfseries SVM};
\end{scope}
\begin{scope}
\path[clip] (  0.00,  0.00) rectangle (505.89,505.89);
\definecolor[named]{drawColor}{rgb}{0.00,0.00,0.00}

\path[draw=drawColor,line width= 0.4pt,line join=round,line cap=round] ( 32.47,377.65) --
	(236.31,377.65) --
	(236.31,473.42) --
	( 32.47,473.42) --
	( 32.47,377.65);
\end{scope}
\begin{scope}
\path[clip] ( 32.47,377.65) rectangle (236.31,473.42);
\definecolor[named]{drawColor}{rgb}{0.00,0.00,1.00}

\path[draw=drawColor,line width= 0.4pt,line join=round,line cap=round] ( 32.47,430.29) -- (236.31,430.29);
\end{scope}
\begin{scope}
\path[clip] (285.42,377.65) rectangle (489.26,473.42);
\definecolor[named]{fillColor}{rgb}{0.75,0.75,0.75}

\path[fill=fillColor] (299.26,425.98) --
	(349.59,425.98) --
	(349.59,460.12) --
	(299.26,460.12) --
	cycle;
\definecolor[named]{drawColor}{rgb}{0.00,0.00,0.00}

\path[draw=drawColor,line width= 1.2pt,line join=round] (299.26,433.07) -- (349.59,433.07);

\path[draw=drawColor,line width= 0.4pt,dash pattern=on 4pt off 4pt ,line join=round,line cap=round] (324.42,421.10) -- (324.42,425.98);

\path[draw=drawColor,line width= 0.4pt,dash pattern=on 4pt off 4pt ,line join=round,line cap=round] (324.42,469.87) -- (324.42,460.12);

\path[draw=drawColor,line width= 0.4pt,line join=round,line cap=round] (311.84,421.10) -- (337.01,421.10);

\path[draw=drawColor,line width= 0.4pt,line join=round,line cap=round] (311.84,469.87) -- (337.01,469.87);

\path[draw=drawColor,line width= 0.4pt,line join=round,line cap=round] (299.26,425.98) --
	(349.59,425.98) --
	(349.59,460.12) --
	(299.26,460.12) --
	(299.26,425.98);

\path[fill=fillColor] (362.17,424.43) --
	(412.50,424.43) --
	(412.50,425.54) --
	(362.17,425.54) --
	cycle;

\path[draw=drawColor,line width= 1.2pt,line join=round] (362.17,425.31) -- (412.50,425.31);

\path[draw=drawColor,line width= 0.4pt,dash pattern=on 4pt off 4pt ,line join=round,line cap=round] (387.34,424.20) -- (387.34,424.43);

\path[draw=drawColor,line width= 0.4pt,dash pattern=on 4pt off 4pt ,line join=round,line cap=round] (387.34,425.54) -- (387.34,425.54);

\path[draw=drawColor,line width= 0.4pt,line join=round,line cap=round] (374.75,424.20) -- (399.92,424.20);

\path[draw=drawColor,line width= 0.4pt,line join=round,line cap=round] (374.75,425.54) -- (399.92,425.54);

\path[draw=drawColor,line width= 0.4pt,line join=round,line cap=round] (362.17,424.43) --
	(412.50,424.43) --
	(412.50,425.54) --
	(362.17,425.54) --
	(362.17,424.43);
\definecolor[named]{fillColor}{rgb}{0.00,1.00,0.00}

\path[fill=fillColor] (425.09,392.50) --
	(475.42,392.50) --
	(475.42,424.43) --
	(425.09,424.43) --
	cycle;

\path[draw=drawColor,line width= 1.2pt,line join=round] (425.09,413.34) -- (475.42,413.34);

\path[draw=drawColor,line width= 0.4pt,dash pattern=on 4pt off 4pt ,line join=round,line cap=round] (450.25,381.20) -- (450.25,392.50);

\path[draw=drawColor,line width= 0.4pt,dash pattern=on 4pt off 4pt ,line join=round,line cap=round] (450.25,428.64) -- (450.25,424.43);

\path[draw=drawColor,line width= 0.4pt,line join=round,line cap=round] (437.67,381.20) -- (462.83,381.20);

\path[draw=drawColor,line width= 0.4pt,line join=round,line cap=round] (437.67,428.64) -- (462.83,428.64);

\path[draw=drawColor,line width= 0.4pt,line join=round,line cap=round] (425.09,392.50) --
	(475.42,392.50) --
	(475.42,424.43) --
	(425.09,424.43) --
	(425.09,392.50);
\end{scope}
\begin{scope}
\path[clip] (  0.00,  0.00) rectangle (505.89,505.89);
\definecolor[named]{drawColor}{rgb}{0.00,0.00,0.00}

\path[draw=drawColor,line width= 0.4pt,line join=round,line cap=round] (324.42,377.65) -- (450.25,377.65);

\path[draw=drawColor,line width= 0.4pt,line join=round,line cap=round] (324.42,377.65) -- (324.42,373.69);

\path[draw=drawColor,line width= 0.4pt,line join=round,line cap=round] (387.34,377.65) -- (387.34,373.69);

\path[draw=drawColor,line width= 0.4pt,line join=round,line cap=round] (450.25,377.65) -- (450.25,373.69);

\node[text=drawColor,anchor=base,inner sep=0pt, outer sep=0pt, scale=  1.3] at (324.42,363.40) {CO - BE};

\node[text=drawColor,anchor=base,inner sep=0pt, outer sep=0pt, scale=  1.3] at (387.34,363.40) {FS - BE};

\node[text=drawColor,anchor=base,inner sep=0pt, outer sep=0pt, scale=  1.3] at (450.25,363.40) {FS - CO};

\path[draw=drawColor,line width= 0.4pt,line join=round,line cap=round] (285.42,381.20) -- (285.42,469.87);

\path[draw=drawColor,line width= 0.4pt,line join=round,line cap=round] (285.42,381.20) -- (281.46,381.20);

\path[draw=drawColor,line width= 0.4pt,line join=round,line cap=round] (285.42,403.37) -- (281.46,403.37);

\path[draw=drawColor,line width= 0.4pt,line join=round,line cap=round] (285.42,425.54) -- (281.46,425.54);

\path[draw=drawColor,line width= 0.4pt,line join=round,line cap=round] (285.42,447.70) -- (281.46,447.70);

\path[draw=drawColor,line width= 0.4pt,line join=round,line cap=round] (285.42,469.87) -- (281.46,469.87);

\node[text=drawColor,rotate= 90.00,anchor=base,inner sep=0pt, outer sep=0pt, scale=  1.2] at (275.91,381.20) {-10};

\node[text=drawColor,rotate= 90.00,anchor=base,inner sep=0pt, outer sep=0pt, scale=  1.2] at (275.91,403.37) {-5};

\node[text=drawColor,rotate= 90.00,anchor=base,inner sep=0pt, outer sep=0pt, scale=  1.2] at (275.91,425.54) {0};

\node[text=drawColor,rotate= 90.00,anchor=base,inner sep=0pt, outer sep=0pt, scale=  1.2] at (275.91,447.70) {5};

\node[text=drawColor,rotate= 90.00,anchor=base,inner sep=0pt, outer sep=0pt, scale=  1.2] at (275.91,469.87) {10};
\end{scope}
\begin{scope}
\path[clip] (252.94,337.26) rectangle (505.89,505.89);
\definecolor[named]{drawColor}{rgb}{0.00,0.00,0.00}

\node[text=drawColor,anchor=base,inner sep=0pt, outer sep=0pt, scale=  1.98] at (387.34,482.82) {\bfseries LDA};
\end{scope}
\begin{scope}
\path[clip] (  0.00,  0.00) rectangle (505.89,505.89);
\definecolor[named]{drawColor}{rgb}{0.00,0.00,0.00}

\path[draw=drawColor,line width= 0.4pt,line join=round,line cap=round] (285.42,377.65) --
	(489.26,377.65) --
	(489.26,473.42) --
	(285.42,473.42) --
	(285.42,377.65);
\end{scope}
\begin{scope}
\path[clip] (285.42,377.65) rectangle (489.26,473.42);
\definecolor[named]{drawColor}{rgb}{0.00,0.00,1.00}

\path[draw=drawColor,line width= 0.4pt,line join=round,line cap=round] (285.42,425.54) -- (489.26,425.54);
\end{scope}
\begin{scope}
\path[clip] ( 32.47,209.02) rectangle (236.31,304.79);
\definecolor[named]{fillColor}{rgb}{0.75,0.75,0.75}

\path[fill=fillColor] ( 46.31,256.35) --
	( 96.64,256.35) --
	( 96.64,285.72) --
	( 46.31,285.72) --
	cycle;
\definecolor[named]{drawColor}{rgb}{0.00,0.00,0.00}

\path[draw=drawColor,line width= 1.2pt,line join=round] ( 46.31,267.43) -- ( 96.64,267.43);

\path[draw=drawColor,line width= 0.4pt,dash pattern=on 4pt off 4pt ,line join=round,line cap=round] ( 71.48,236.95) -- ( 71.48,256.35);

\path[draw=drawColor,line width= 0.4pt,dash pattern=on 4pt off 4pt ,line join=round,line cap=round] ( 71.48,301.24) -- ( 71.48,285.72);

\path[draw=drawColor,line width= 0.4pt,line join=round,line cap=round] ( 58.90,236.95) -- ( 84.06,236.95);

\path[draw=drawColor,line width= 0.4pt,line join=round,line cap=round] ( 58.90,301.24) -- ( 84.06,301.24);

\path[draw=drawColor,line width= 0.4pt,line join=round,line cap=round] ( 46.31,256.35) --
	( 96.64,256.35) --
	( 96.64,285.72) --
	( 46.31,285.72) --
	( 46.31,256.35);

\path[fill=fillColor] (109.23,265.77) --
	(159.56,265.77) --
	(159.56,274.09) --
	(109.23,274.09) --
	cycle;

\path[draw=drawColor,line width= 1.2pt,line join=round] (109.23,267.43) -- (159.56,267.43);

\path[draw=drawColor,line width= 0.4pt,dash pattern=on 4pt off 4pt ,line join=round,line cap=round] (134.39,265.77) -- (134.39,265.77);

\path[draw=drawColor,line width= 0.4pt,dash pattern=on 4pt off 4pt ,line join=round,line cap=round] (134.39,275.75) -- (134.39,274.09);

\path[draw=drawColor,line width= 0.4pt,line join=round,line cap=round] (121.81,265.77) -- (146.98,265.77);

\path[draw=drawColor,line width= 0.4pt,line join=round,line cap=round] (121.81,275.75) -- (146.98,275.75);

\path[draw=drawColor,line width= 0.4pt,line join=round,line cap=round] (109.23,265.77) --
	(159.56,265.77) --
	(159.56,274.09) --
	(109.23,274.09) --
	(109.23,265.77);

\path[fill=fillColor] (172.14,236.95) --
	(222.47,236.95) --
	(222.47,276.86) --
	(172.14,276.86) --
	cycle;

\path[draw=drawColor,line width= 1.2pt,line join=round] (172.14,265.77) -- (222.47,265.77);

\path[draw=drawColor,line width= 0.4pt,dash pattern=on 4pt off 4pt ,line join=round,line cap=round] (197.31,212.57) -- (197.31,236.95);

\path[draw=drawColor,line width= 0.4pt,dash pattern=on 4pt off 4pt ,line join=round,line cap=round] (197.31,297.92) -- (197.31,276.86);

\path[draw=drawColor,line width= 0.4pt,line join=round,line cap=round] (184.72,212.57) -- (209.89,212.57);

\path[draw=drawColor,line width= 0.4pt,line join=round,line cap=round] (184.72,297.92) -- (209.89,297.92);

\path[draw=drawColor,line width= 0.4pt,line join=round,line cap=round] (172.14,236.95) --
	(222.47,236.95) --
	(222.47,276.86) --
	(172.14,276.86) --
	(172.14,236.95);
\end{scope}
\begin{scope}
\path[clip] (  0.00,  0.00) rectangle (505.89,505.89);
\definecolor[named]{drawColor}{rgb}{0.00,0.00,0.00}

\path[draw=drawColor,line width= 0.4pt,line join=round,line cap=round] ( 71.48,209.02) -- (197.31,209.02);

\path[draw=drawColor,line width= 0.4pt,line join=round,line cap=round] ( 71.48,209.02) -- ( 71.48,205.06);

\path[draw=drawColor,line width= 0.4pt,line join=round,line cap=round] (134.39,209.02) -- (134.39,205.06);

\path[draw=drawColor,line width= 0.4pt,line join=round,line cap=round] (197.31,209.02) -- (197.31,205.06);

\node[text=drawColor,anchor=base,inner sep=0pt, outer sep=0pt, scale=  1.3] at ( 71.48,194.77) {CO - BE};

\node[text=drawColor,anchor=base,inner sep=0pt, outer sep=0pt, scale=  1.3] at (134.39,194.77) {FS - BE};

\node[text=drawColor,anchor=base,inner sep=0pt, outer sep=0pt, scale=  1.3] at (197.31,194.77) {FS - CO};

\path[draw=drawColor,line width= 0.4pt,line join=round,line cap=round] ( 32.47,221.44) -- ( 32.47,287.94);

\path[draw=drawColor,line width= 0.4pt,line join=round,line cap=round] ( 32.47,221.44) -- ( 28.51,221.44);

\path[draw=drawColor,line width= 0.4pt,line join=round,line cap=round] ( 32.47,243.60) -- ( 28.51,243.60);

\path[draw=drawColor,line width= 0.4pt,line join=round,line cap=round] ( 32.47,265.77) -- ( 28.51,265.77);

\path[draw=drawColor,line width= 0.4pt,line join=round,line cap=round] ( 32.47,287.94) -- ( 28.51,287.94);

\node[text=drawColor,rotate= 90.00,anchor=base,inner sep=0pt, outer sep=0pt, scale=  1.2] at ( 22.97,221.44) {-4};

\node[text=drawColor,rotate= 90.00,anchor=base,inner sep=0pt, outer sep=0pt, scale=  1.2] at ( 22.97,243.60) {-2};

\node[text=drawColor,rotate= 90.00,anchor=base,inner sep=0pt, outer sep=0pt, scale=  1.2] at ( 22.97,265.77) {0};

\node[text=drawColor,rotate= 90.00,anchor=base,inner sep=0pt, outer sep=0pt, scale=  1.2] at ( 22.97,287.94) {2};
\end{scope}
\begin{scope}
\path[clip] (  0.00,168.63) rectangle (252.94,337.26);
\definecolor[named]{drawColor}{rgb}{0.00,0.00,0.00}

\node[text=drawColor,anchor=base,inner sep=0pt, outer sep=0pt, scale=  1.98] at (134.39,314.19) {\bfseries CART};
\end{scope}
\begin{scope}
\path[clip] (  0.00,  0.00) rectangle (505.89,505.89);
\definecolor[named]{drawColor}{rgb}{0.00,0.00,0.00}

\path[draw=drawColor,line width= 0.4pt,line join=round,line cap=round] ( 32.47,209.02) --
	(236.31,209.02) --
	(236.31,304.79) --
	( 32.47,304.79) --
	( 32.47,209.02);
\end{scope}
\begin{scope}
\path[clip] ( 32.47,209.02) rectangle (236.31,304.79);
\definecolor[named]{drawColor}{rgb}{0.00,0.00,1.00}

\path[draw=drawColor,line width= 0.4pt,line join=round,line cap=round] ( 32.47,265.77) -- (236.31,265.77);
\end{scope}
\begin{scope}
\path[clip] (285.42,209.02) rectangle (489.26,304.79);
\definecolor[named]{fillColor}{rgb}{0.75,0.75,0.75}

\path[fill=fillColor] (299.26,249.06) --
	(349.59,249.06) --
	(349.59,270.25) --
	(299.26,270.25) --
	cycle;
\definecolor[named]{drawColor}{rgb}{0.00,0.00,0.00}

\path[draw=drawColor,line width= 1.2pt,line join=round] (299.26,251.80) -- (349.59,251.80);

\path[draw=drawColor,line width= 0.4pt,dash pattern=on 4pt off 4pt ,line join=round,line cap=round] (324.42,246.31) -- (324.42,249.06);

\path[draw=drawColor,line width= 0.4pt,dash pattern=on 4pt off 4pt ,line join=round,line cap=round] (324.42,277.70) -- (324.42,270.25);

\path[draw=drawColor,line width= 0.4pt,line join=round,line cap=round] (311.84,246.31) -- (337.01,246.31);

\path[draw=drawColor,line width= 0.4pt,line join=round,line cap=round] (311.84,277.70) -- (337.01,277.70);

\path[draw=drawColor,line width= 0.4pt,line join=round,line cap=round] (299.26,249.06) --
	(349.59,249.06) --
	(349.59,270.25) --
	(299.26,270.25) --
	(299.26,249.06);

\path[fill=fillColor] (362.17,238.86) --
	(412.50,238.86) --
	(412.50,253.37) --
	(362.17,253.37) --
	cycle;

\path[draw=drawColor,line width= 1.2pt,line join=round] (362.17,244.74) -- (412.50,244.74);

\path[draw=drawColor,line width= 0.4pt,dash pattern=on 4pt off 4pt ,line join=round,line cap=round] (387.34,227.48) -- (387.34,238.86);

\path[draw=drawColor,line width= 0.4pt,dash pattern=on 4pt off 4pt ,line join=round,line cap=round] (387.34,258.87) -- (387.34,253.37);

\path[draw=drawColor,line width= 0.4pt,line join=round,line cap=round] (374.75,227.48) -- (399.92,227.48);

\path[draw=drawColor,line width= 0.4pt,line join=round,line cap=round] (374.75,258.87) -- (399.92,258.87);

\path[draw=drawColor,line width= 0.4pt,line join=round,line cap=round] (362.17,238.86) --
	(412.50,238.86) --
	(412.50,253.37) --
	(362.17,253.37) --
	(362.17,238.86);
\definecolor[named]{fillColor}{rgb}{0.00,1.00,0.00}

\path[fill=fillColor] (425.09,220.02) --
	(475.42,220.02) --
	(475.42,247.49) --
	(425.09,247.49) --
	cycle;

\path[draw=drawColor,line width= 1.2pt,line join=round] (425.09,241.60) -- (475.42,241.60);

\path[draw=drawColor,line width= 0.4pt,dash pattern=on 4pt off 4pt ,line join=round,line cap=round] (450.25,212.57) -- (450.25,220.02);

\path[draw=drawColor,line width= 0.4pt,dash pattern=on 4pt off 4pt ,line join=round,line cap=round] (450.25,301.24) -- (450.25,247.49);

\path[draw=drawColor,line width= 0.4pt,line join=round,line cap=round] (437.67,212.57) -- (462.83,212.57);

\path[draw=drawColor,line width= 0.4pt,line join=round,line cap=round] (437.67,301.24) -- (462.83,301.24);

\path[draw=drawColor,line width= 0.4pt,line join=round,line cap=round] (425.09,220.02) --
	(475.42,220.02) --
	(475.42,247.49) --
	(425.09,247.49) --
	(425.09,220.02);
\end{scope}
\begin{scope}
\path[clip] (  0.00,  0.00) rectangle (505.89,505.89);
\definecolor[named]{drawColor}{rgb}{0.00,0.00,0.00}

\path[draw=drawColor,line width= 0.4pt,line join=round,line cap=round] (324.42,209.02) -- (450.25,209.02);

\path[draw=drawColor,line width= 0.4pt,line join=round,line cap=round] (324.42,209.02) -- (324.42,205.06);

\path[draw=drawColor,line width= 0.4pt,line join=round,line cap=round] (387.34,209.02) -- (387.34,205.06);

\path[draw=drawColor,line width= 0.4pt,line join=round,line cap=round] (450.25,209.02) -- (450.25,205.06);

\node[text=drawColor,anchor=base,inner sep=0pt, outer sep=0pt, scale=  1.3] at (324.42,194.77) {CO - BE};

\node[text=drawColor,anchor=base,inner sep=0pt, outer sep=0pt, scale=  1.3] at (387.34,194.77) {FS - BE};

\node[text=drawColor,anchor=base,inner sep=0pt, outer sep=0pt, scale=  1.3] at (450.25,194.77) {FS - CO};

\path[draw=drawColor,line width= 0.4pt,line join=round,line cap=round] (285.42,220.42) -- (285.42,298.89);

\path[draw=drawColor,line width= 0.4pt,line join=round,line cap=round] (285.42,220.42) -- (281.46,220.42);

\path[draw=drawColor,line width= 0.4pt,line join=round,line cap=round] (285.42,236.11) -- (281.46,236.11);

\path[draw=drawColor,line width= 0.4pt,line join=round,line cap=round] (285.42,251.80) -- (281.46,251.80);

\path[draw=drawColor,line width= 0.4pt,line join=round,line cap=round] (285.42,267.50) -- (281.46,267.50);

\path[draw=drawColor,line width= 0.4pt,line join=round,line cap=round] (285.42,283.19) -- (281.46,283.19);

\path[draw=drawColor,line width= 0.4pt,line join=round,line cap=round] (285.42,298.89) -- (281.46,298.89);

\node[text=drawColor,rotate= 90.00,anchor=base,inner sep=0pt, outer sep=0pt, scale=  1.2] at (275.91,220.42) {-4};

\node[text=drawColor,rotate= 90.00,anchor=base,inner sep=0pt, outer sep=0pt, scale=  1.2] at (275.91,251.80) {0};

\node[text=drawColor,rotate= 90.00,anchor=base,inner sep=0pt, outer sep=0pt, scale=  1.2] at (275.91,267.50) {2};

\node[text=drawColor,rotate= 90.00,anchor=base,inner sep=0pt, outer sep=0pt, scale=  1.2] at (275.91,283.19) {4};

\node[text=drawColor,rotate= 90.00,anchor=base,inner sep=0pt, outer sep=0pt, scale=  1.2] at (275.91,298.89) {6};
\end{scope}
\begin{scope}
\path[clip] (252.94,168.63) rectangle (505.89,337.26);
\definecolor[named]{drawColor}{rgb}{0.00,0.00,0.00}

\node[text=drawColor,anchor=base,inner sep=0pt, outer sep=0pt, scale=  1.98] at (387.34,314.19) {\bfseries RF};
\end{scope}
\begin{scope}
\path[clip] (  0.00,  0.00) rectangle (505.89,505.89);
\definecolor[named]{drawColor}{rgb}{0.00,0.00,0.00}

\path[draw=drawColor,line width= 0.4pt,line join=round,line cap=round] (285.42,209.02) --
	(489.26,209.02) --
	(489.26,304.79) --
	(285.42,304.79) --
	(285.42,209.02);
\end{scope}
\begin{scope}
\path[clip] (285.42,209.02) rectangle (489.26,304.79);
\definecolor[named]{drawColor}{rgb}{0.00,0.00,1.00}

\path[draw=drawColor,line width= 0.4pt,line join=round,line cap=round] (285.42,251.80) -- (489.26,251.80);
\end{scope}
\begin{scope}
\path[clip] ( 32.47, 40.39) rectangle (236.31,136.16);
\definecolor[named]{fillColor}{rgb}{0.75,0.75,0.75}

\path[fill=fillColor] ( 46.31, 73.69) --
	( 96.64, 73.69) --
	( 96.64, 98.64) --
	( 46.31, 98.64) --
	cycle;
\definecolor[named]{drawColor}{rgb}{0.00,0.00,0.00}

\path[draw=drawColor,line width= 1.2pt,line join=round] ( 46.31, 85.20) -- ( 96.64, 85.20);

\path[draw=drawColor,line width= 0.4pt,dash pattern=on 4pt off 4pt ,line join=round,line cap=round] ( 71.48, 68.12) -- ( 71.48, 73.69);

\path[draw=drawColor,line width= 0.4pt,dash pattern=on 4pt off 4pt ,line join=round,line cap=round] ( 71.48,132.61) -- ( 71.48, 98.64);

\path[draw=drawColor,line width= 0.4pt,line join=round,line cap=round] ( 58.90, 68.12) -- ( 84.06, 68.12);

\path[draw=drawColor,line width= 0.4pt,line join=round,line cap=round] ( 58.90,132.61) -- ( 84.06,132.61);

\path[draw=drawColor,line width= 0.4pt,line join=round,line cap=round] ( 46.31, 73.69) --
	( 96.64, 73.69) --
	( 96.64, 98.64) --
	( 46.31, 98.64) --
	( 46.31, 73.69);

\path[fill=fillColor] (109.23, 71.00) --
	(159.56, 71.00) --
	(159.56, 83.09) --
	(109.23, 83.09) --
	cycle;

\path[draw=drawColor,line width= 1.2pt,line join=round] (109.23, 79.25) -- (159.56, 79.25);

\path[draw=drawColor,line width= 0.4pt,dash pattern=on 4pt off 4pt ,line join=round,line cap=round] (134.39, 54.69) -- (134.39, 71.00);

\path[draw=drawColor,line width= 0.4pt,dash pattern=on 4pt off 4pt ,line join=round,line cap=round] (134.39, 96.14) -- (134.39, 83.09);

\path[draw=drawColor,line width= 0.4pt,line join=round,line cap=round] (121.81, 54.69) -- (146.98, 54.69);

\path[draw=drawColor,line width= 0.4pt,line join=round,line cap=round] (121.81, 96.14) -- (146.98, 96.14);

\path[draw=drawColor,line width= 0.4pt,line join=round,line cap=round] (109.23, 71.00) --
	(159.56, 71.00) --
	(159.56, 83.09) --
	(109.23, 83.09) --
	(109.23, 71.00);

\path[fill=fillColor] (172.14, 56.03) --
	(222.47, 56.03) --
	(222.47, 80.60) --
	(172.14, 80.60) --
	cycle;

\path[draw=drawColor,line width= 1.2pt,line join=round] (172.14, 68.31) -- (222.47, 68.31);

\path[draw=drawColor,line width= 0.4pt,dash pattern=on 4pt off 4pt ,line join=round,line cap=round] (197.31, 43.94) -- (197.31, 56.03);

\path[draw=drawColor,line width= 0.4pt,dash pattern=on 4pt off 4pt ,line join=round,line cap=round] (197.31, 91.15) -- (197.31, 80.60);

\path[draw=drawColor,line width= 0.4pt,line join=round,line cap=round] (184.72, 43.94) -- (209.89, 43.94);

\path[draw=drawColor,line width= 0.4pt,line join=round,line cap=round] (184.72, 91.15) -- (209.89, 91.15);

\path[draw=drawColor,line width= 0.4pt,line join=round,line cap=round] (172.14, 56.03) --
	(222.47, 56.03) --
	(222.47, 80.60) --
	(172.14, 80.60) --
	(172.14, 56.03);
\end{scope}
\begin{scope}
\path[clip] (  0.00,  0.00) rectangle (505.89,505.89);
\definecolor[named]{drawColor}{rgb}{0.00,0.00,0.00}

\path[draw=drawColor,line width= 0.4pt,line join=round,line cap=round] ( 71.48, 40.39) -- (197.31, 40.39);

\path[draw=drawColor,line width= 0.4pt,line join=round,line cap=round] ( 71.48, 40.39) -- ( 71.48, 36.43);

\path[draw=drawColor,line width= 0.4pt,line join=round,line cap=round] (134.39, 40.39) -- (134.39, 36.43);

\path[draw=drawColor,line width= 0.4pt,line join=round,line cap=round] (197.31, 40.39) -- (197.31, 36.43);

\node[text=drawColor,anchor=base,inner sep=0pt, outer sep=0pt, scale=  1.3] at ( 71.48, 26.14) {CO - BE};

\node[text=drawColor,anchor=base,inner sep=0pt, outer sep=0pt, scale=  1.3] at (134.39, 26.14) {FS - BE};

\node[text=drawColor,anchor=base,inner sep=0pt, outer sep=0pt, scale=  1.3] at (197.31, 26.14) {FS - CO};

\path[draw=drawColor,line width= 0.4pt,line join=round,line cap=round] ( 32.47, 42.02) -- ( 32.47,118.79);

\path[draw=drawColor,line width= 0.4pt,line join=round,line cap=round] ( 32.47, 42.02) -- ( 28.51, 42.02);

\path[draw=drawColor,line width= 0.4pt,line join=round,line cap=round] ( 32.47, 61.21) -- ( 28.51, 61.21);

\path[draw=drawColor,line width= 0.4pt,line join=round,line cap=round] ( 32.47, 80.41) -- ( 28.51, 80.41);

\path[draw=drawColor,line width= 0.4pt,line join=round,line cap=round] ( 32.47, 99.60) -- ( 28.51, 99.60);

\path[draw=drawColor,line width= 0.4pt,line join=round,line cap=round] ( 32.47,118.79) -- ( 28.51,118.79);

\node[text=drawColor,rotate= 90.00,anchor=base,inner sep=0pt, outer sep=0pt, scale=  1.2] at ( 22.97, 42.02) {-10};

\node[text=drawColor,rotate= 90.00,anchor=base,inner sep=0pt, outer sep=0pt, scale=  1.2] at ( 22.97, 80.41) {0};

\node[text=drawColor,rotate= 90.00,anchor=base,inner sep=0pt, outer sep=0pt, scale=  1.2] at ( 22.97, 99.60) {5};

\node[text=drawColor,rotate= 90.00,anchor=base,inner sep=0pt, outer sep=0pt, scale=  1.2] at ( 22.97,118.79) {10};
\end{scope}
\begin{scope}
\path[clip] (  0.00,  0.00) rectangle (252.94,168.63);
\definecolor[named]{drawColor}{rgb}{0.00,0.00,0.00}

\node[text=drawColor,anchor=base,inner sep=0pt, outer sep=0pt, scale=  1.98] at (134.39,145.56) {\bfseries NN};
\end{scope}
\begin{scope}
\path[clip] (  0.00,  0.00) rectangle (505.89,505.89);
\definecolor[named]{drawColor}{rgb}{0.00,0.00,0.00}

\path[draw=drawColor,line width= 0.4pt,line join=round,line cap=round] ( 32.47, 40.39) --
	(236.31, 40.39) --
	(236.31,136.16) --
	( 32.47,136.16) --
	( 32.47, 40.39);
\end{scope}
\begin{scope}
\path[clip] ( 32.47, 40.39) rectangle (236.31,136.16);
\definecolor[named]{drawColor}{rgb}{0.00,0.00,1.00}

\path[draw=drawColor,line width= 0.4pt,line join=round,line cap=round] ( 32.47, 80.41) -- (236.31, 80.41);
\end{scope}
\end{tikzpicture}

   }
\caption{Comparing the search strategies for mRMR. Results of the post-hoc tests for each classifier.}
\label{mrmr-cl}
\end{figure}
  The second interesting point is with respect to the Madelon dataset. As can be seen, mRMR with greedy search algorithms perform poorly on this dataset. Several authors have already utilized this dataset to compare their proposed criterion with mRMR and arrived at the conclusion that mRMR cannot handle highly correlated features, as in Madelon dataset. However, surprisingly the performance of the mRMR+COBRA is as good as JMI on this dataset meaning that it is not the criterion but the search method that has difficulty to deal with highly correlated  features. Thus, any conclusion with respect to the quality of a measure has to be drawn carefully since, as in this case, the effect of the non optimum search method can be decisive.
   
  To discover the statistically meaningful differences between the algorithms, we applied the Friedman test following with Wilcoxon-Nemenyi post-hoc analysis, as suggested in \cite{hollander:99}, on the average accuracies (the last column of Table \ref{ssp-be}). Note that since we have 8 datasets, there are 8 independent measurements available for each algorithm. The results of this test for mRMR based algorithms have been depicted in Figure \ref{mrmr-mean}. In all box plots, CO stands for COBRA algorithm. Each box plot compares a pair of the algorithms. The green box plots represent the existence of a significant difference between the corresponding algorithms. The adjusted p-values for each pair of algorithms have also been reported in Figure \ref{mrmr-mean}. The smaller the p-value, the stronger the evidence against the null hypothesis. As can be seen, COBRA shows meaningful superiority over both greedy algorithms. However, if we set the significance level at $p=0.05$, only FS rejects the null 
hypothesis and shows a meaningful difference with COBRA.
  
  The same test was run for each classifier and its results can be found in Figure \ref{mrmr-cl}. While three of the classifiers show some differences between FS and COBRA, neither of them reveal any meaningful difference between BE and COBRA. At this point, the least we can conclude is that independent of the classification algorithm we choose, it is a good chance that FS performs worse than COBRA. 
  
  For JMI, however, the performances of all algorithms are comparable and with only 8 datasets it is difficult to draw any conclusion. Thus, the Wilcoxon-Nemenyi test results for JMI is not shown here because of the lack of space.
 \vspace{-10mm}
    
\begin{center}
  \begin{table*}
    \addtolength{\tabcolsep}{4.4mm}
    \hfill{}
    \begin{tabular}{l|| c c c c c} 
      \hline \Ia
   \textbf{Datasets}                     & \bf{MAD}         & \bf{NCI}              &     \bf{IAD}               &  \bf{ARR}             &  \bf{CNA}                         \\ \hline\hline \Ib 
   \bf{mRMR+COBRA}                       &  74.81$\pmS$0.65  & 73.67$\pmS$2.41        &      96.64$\pmS$0.16        &   77.75$\pmS$1.03      & 88.91$\pmS$0.31                      \\  
   \bf{mRMR+QPFS}                        &  71.44$\pmS$0.57  & 71.00$\pmS$1.84        &      95.02$\pmS$0.21        &   78.73$\pmS$0.84      & 86.93$\pmS$0.45                       \\ 
   \bf{mRMR+SOSS}                        &  71.36$\pmS$0.53  & 72.65$\pmS$2.13        &      96.64$\pmS$0.28        &   79.86$\pmS$1.18      & 85.43$\pmS$0.49                        \\  \hline \hline \Ia
                                       
   \bf{Time COBRA}                       &  175${\,+\,}$24  &368${\,+\,}$341          &     540${\,+\,}$121        &  6${\,+\,}$14          & 120${\,+\,}$50                      \\  
   \bf{Time QPFS}                        &    11            &  180                    &      202                   &    1                   & 25                       \\ 
   \bf{Time SOSS}                        &  175${\,+\,}$5   &  368${\,+\,}$27         &     540${\,+\,}$12          &  6${\,+\,}$4          & 120${\,+\,}$7                        \\  \hline
                                             
      \end{tabular}
      \hfill{}
       \caption{Comparison of COBRA with QPFS and SOSS over 5 datasets. Average classification rates and their standard deviations are reported in the top three rows of the table. In the next three rows, the computational times in second are shown where the first value for COBRA and SOSS is for calculating the mutual information matrix and the second value is the time needed to solve the optimization problems.}
       \label{qpfs}
  \end{table*} 
 \end{center}
  In the next experiment COBRA is compared with two other convex programming based feature selection algorithms, SOSS \cite{naghibi:13} and QPFS \cite{rod:10}. Both SOSS and QPFS employ quadratic programing techniques to maximize a score function. SOSS, however, uses an instance of randomized rounding to generate the set-membership binary values while QPFS ranks the features based on their scores (achieved from solving the convex problem) and therefore, sidesteps the difficulties of generating binary values. Note that both COBRA and SOSS first need to calculate the mutual information matrix $\mathbf{Q}$. Once it is calculated, they can solve their corresponding convex optimization problems for different values of $P$. The first 3 rows of Table \ref{qpfs} report the average (over 5 classifiers) classification accuracies of these three algorithms and the standard deviation of these mean accuracies (calculated over the cross-validation folds). In the next three rows of the table, the computational times of each 
algorithm for a single run (in second) are shown, i.e., the amount of time needed to select a feature set with (given) $P$ features. The reported times for COBRA and SOSS consist of two values. The first value is the time needed to calculate the mutual information matrix $\mathbf{Q}$ and the second value is the amount of time needed to solve the corresponding convex optimization problem. All the values were measured on a PC with an Intel Core i7 CPU. As seen, QPFS is significantly faster than COBRA and SOSS. This computational superiority, however, seems to come at the expense of lower classification accuracy. For large datasets such as IAD, CNA and MAD, the  Nystr\"{o}m approximation used in QPFS to cast the problem into a lower dimensional subspace does not yield a precise enough approximation and results in lower classification accuracies. An important remark to interpret these results is that, for NCI dataset (in all the experiments) we first filtered out the features with the low mutual information 
values with the class label and only kept 2000 informative features (similarly for DEX and DBW datasets). Thus, the dimension is 2000 and not 9703 as mentioned in Table \ref{datasets}. 

  The generalization power of the COBRA algorithm over different classifiers is another important issue to test. As can be observed in Table \ref{ssp-be}, the number of selected features varies quite markedly from one classifier to another. However, based on our experiments, the optimum feature set of any of the classifiers, usually (for large enough datasets) achieves a near-optimal accuracy in conjunction with other classifiers as well. This is shown in Table \ref{general} for 4 classifiers and 3 datasets. The COBRA features of the LDA classifier in Table \ref{ssp-be} is used here to train other classifiers. Table \ref{general} lists the accuracies obtained by using the LDA features and the optimal features, repeated from Table \ref{ssp-be}. Unlike the CNA and  IAD datasets, a significant accuracy reduction can be observed in the case of ARR data which has substantially less training data than CNA and IAD. It suggests that for small size datasets, a feature 
selection scheme should take the induction algorithm into account since the learning algorithm is sensitive to small changes of the feature set.       
\begin{table}[ht]
  \addtolength{\tabcolsep}{1.3mm}

\begin{tabular}{l|c||c c c c } 
\hline 
  \multicolumn{2}{c||}{\bf{Classifiers}}            & \textbf{SVM}         & \textbf{CART}                   & \textbf{RF}          & \textbf{NN}              \Ia          \\ \hline\hline
                                 
   \multirow{2}{*}{\textbf{ARR}}       &   LDA feat.     &78.4         & 73.7                   & 77.1        & 68.00                           \\  \cline{2-2}
                              &   Optimum       &81.9         & 75.4                   & 82.2        & 72.9                                        \\  \hline\hline
  \multirow{2}{*}{\textbf{CNA}}        &  LDA feat.      & 92.6        & 75.0                   & 90.5        & 91.1                       \\  \cline{2-2}
                              & Optimum         & 94.0        & 75.0                   & 90.8        & 92.0                        \\  \hline\hline
                              
  \multirow{2}{*}{\textbf{IAD}}        &  LDA feat      & 95.8        & 96.0                   & 97.2        & 96.3                          \\  \cline{2-2}
                              &  Optimum       & 96.5        & 96.4                   & 97.2        & 97.1                           \\  \hline

   \end{tabular}
\caption{The performance of the classification algorithms when trained with COBRA features optimized for the LDA classifier. This table shows the generalization power of the COBRA features on the classifiers.}
\label{general}
\end{table}
\vspace{-5mm}
\section{Conclusion}
\label{con}
A convex based parallel search strategy for feature selection, COBRA, was suggested in this work. Its approximation ratio was derived and compared with the approximation ratio of the backward elimination method. It was experimentally shown that COBRA outperforms sequential search methods especially in the case of sparse data.  Moreover, we presented two series expansions for mutual information, and showed that most mutual information based score functions in the literature including mRMR and MIFS are truncated approximations of these expansions. Furthermore, the underlying connection between MIFS and the Kirwood approximation was explored, and it was shown that by adopting the class conditional independence assumption and the Kirkwood approximation for $Pr(\mathbf{X})$, mutual information reduces to the MIFS criterion. 
\section{Acknowledgments}
This work has partly been supported by Swiss National Science Foundation (SNSF).

\bibliographystyle{plain}

\begin{thebibliography}{10}

\bibitem{abramson:63}
N.~Abramson.
\newblock {\em Information theory and coding}.
\newblock {McGraw-Hill}, New York, 1963.

\bibitem{aha:94}
D.~W. Aha and R.~L. Bankert.
\newblock A comparative evaluation of sequential feature selection algorithms.
\newblock In {\em Learning from Data: Artificial Intelligence and Statistics
  V}. Springer-Verlag, 1996.

\bibitem{asahiro:00}
Y.~Asahiro, K.~Iwama, H.~Tamaki, and T.~Tokuyama.
\newblock Greedily finding a dense subgraph.
\newblock {\em Journal of Algorithms}, 2000.

\bibitem{balagani:10}
K.~S. Balagani and V.~V. Phoha.
\newblock On the feature selection criterion based on an approximation of
  multidimensional mutual information.
\newblock {\em IEEE Trans. on Pattern Analysis and Machine Intelligence}, 2010.

\bibitem{battiti:94}
R.~Battiti.
\newblock Using mutual information for selecting features in supervised neural
  net learning.
\newblock {\em IEEE Trans. on Neural Networks}, 5:537--550, 1994.

\bibitem{bog:07}
V.~I. Bogachev.
\newblock {\em Measure {T}heory}.
\newblock Springer, 2007.

\bibitem{boyd:04}
S.~Boyd and L.~Vandenberghe.
\newblock {\em Convex Optimization}.
\newblock Cambridge University Press, New York, NY, USA, 2004.

\bibitem{brown:09}
G.~Brown.
\newblock A new perspective for information theoretic feature selection.
\newblock In {\em Proceedings of Artificial Intelligence and Statistics}, 2009.

\bibitem{brown:12}
G.~Brown, A.~Pocock, M.~Zhao, and M.~Luj\'{a}n.
\newblock Conditional likelihood maximisation: A unifying framework for
  information theoretic feature selection.
\newblock {\em Journal of Machine Learning}, 2012.

\bibitem{ciarelli:10}
P.~M. Ciarelli, E.~O.~T. Salles, and E.~Oliveira.
\newblock An evolving system based on probabilistic neural network.
\newblock In {\em Proceedings of BSNN}, 2010.

\bibitem{cover:91}
T.~M. Cover and J.~A. Thomas.
\newblock {\em Elements of Information Theory}.
\newblock Wiley-Interscience, New York, NY, USA, 1991.

\bibitem{doak:92}
J.~Doak.
\newblock An evaluation of feature selection methods and their application to
  computer security.
\newblock Technical Report CSE-92-18, University of California at Davis, 1992.

\bibitem{fano:61}
R.~Fano.
\newblock {\em Transmission of Information: A Statistical Theory of
  Communications}.
\newblock The MIT Press, Cambridge, MA, 1961.

\bibitem{frank:10}
A.~Frank and A.~Asuncion.
\newblock {UCI} machine learning repository, 2010.

\bibitem{friedman:74}
J.~H. Friedman and J.~W. Tukey.
\newblock A projection pursuit algorithm for exploratory data analysis.
\newblock {\em IEEE Trans. on Computers}, 1974.

\bibitem{frieze:83}
A.~M. Frieze and J.~Yadegar.
\newblock On the quadratic assignment problem.
\newblock {\em Discrete Applied Mathematics}, 1983.

\bibitem{ben:12}
B.~Frénay, G.~Doquire, and M.~Verleysen.
\newblock On the potential inadequacy of mutual information for feature
  selection.
\newblock In {\em Proceedings of ESANN}, 2012.

\bibitem{goemans:95}
M.~X. Goemans and D.~P. Williamson.
\newblock Improved approximation algorithms for maximum cut and satisfiability
  problems using semidefinite programming.
\newblock {\em Journal of the ACM}, 1995.

\bibitem{grippo:12}
L.~Grippo, L.~Palagi, M.~Piacentini, V.~Piccialli, and G.~Rinaldi.
\newblock {SpeeDP}: an algorithm to compute {SDP} bounds for very large max-cut
  instances.
\newblock {\em Mathematical Programming}, 2012.

\bibitem{gurban:09}
M.~Gurban.
\newblock {\em Multimodal feature extraction and fusion for audio-visual speech
  recognition}.
\newblock PhD thesis, {4292}, STI, EPF Lausanne, 2009.

\bibitem{han:80}
T.~S. Han.
\newblock Multiple mutual informations and multiple interactions in frequency
  data.
\newblock {\em Information and Control}, 1980.

\bibitem{hellman:70}
M.~Hellman and J.~Raviv.
\newblock Probability of error, equivocation and the {C}hernoff bound.
\newblock {\em IEEE Trans. on Information Theory}, 1970.

\bibitem{hollander:99}
M.~Hollander and D.~A. Wolfe.
\newblock {\em Nonparametric Statistical Methods, 2nd Edition}.
\newblock Wiley-Interscience, 1999.

\bibitem{bao:11}
B.~G. Hu.
\newblock What are the differences between {Bayesian} classifiers and
  mutual-information classifiers?
\newblock {\em {IEEE} Trans. on Neural Networks and Learning Systems}, 2011.

\bibitem{karger:01}
D.~Karger and N.~Srebro.
\newblock Learning {M}arkov networks: Maximum bounded tree-width graphs.
\newblock {\em Proceedings of the 12th Annual Symposium on Discrete
  Algorithms}, pages 392--401, 2001.

\bibitem{killian:07}
B.~J. Killian, J.~Y. Kravitz, and M.~K. Gilson.
\newblock {Extraction of configurational entropy from molecular simulations via
  an expansion approximation}.
\newblock {\em Journal of Chemical Physics}, 2007.

\bibitem{kohavi:96}
R.~Kohavi.
\newblock {\em Wrappers for performance enhancement and oblivious decision
  graphs}.
\newblock PhD thesis, Stanford, CA, USA, 1996.

\bibitem{kwak:02}
N.~Kwak and C.~Choi.
\newblock Input feature selection for classification problems.
\newblock {\em IEEE Trans. on Neural Networks}, 2002.

\bibitem{lof:04}
J.~Lofberg.
\newblock Yalmip: a toolbox for modeling and optimization in matlab.
\newblock In {\em Proceedings of Computer Aided Control Systems Design}, 2004.

\bibitem{mcgill:54}
W.~McGill.
\newblock Multivariate information transmission.
\newblock {\em IRE Professional Group on Information Theory}, 1954.

\bibitem{meyer:06}
P.~Meyer and G.~Bontempi.
\newblock {\em On the Use of Variable Complementarity for Feature Selection in
  Cancer Classification}.
\newblock Springer, 2006.

\bibitem{naghibi:13}
T.~Naghibi, S.~Hoffmann, and B.~Pfister.
\newblock Convex approximation of the {NP}-hard search problem in feature
  subset selection.
\newblock In {\em Proceedings of ICASSP}, 2013.

\bibitem{narendra:77}
P.~M. Narendra and K.~Fukunaga.
\newblock A branch and bound algorithm for feature subset selection.
\newblock {\em IEEE Trans. on Computers}, 1977.

\bibitem{neumann:04}
J.~Neumann, C.~Schnörr, and G.~Steidl.
\newblock {SVM}-based feature selection by direct objective minimisation.
\newblock In {\em Proceedings of DAGM}, 2004.

\bibitem{peng:05}
H.~Peng, F.~Long, and C.~Ding.
\newblock Feature selection based on mutual information: criteria of
  max-dependency, max-relevance, and min-redundancy.
\newblock {\em IEEE Trans. on Pattern Analysis and Machine Intelligence}, 2005.

\bibitem{poljak:95}
S.~Poljak, F.~Rendl, and H.~Wolkowicz.
\newblock A recipe for semidefinite relaxation for (0,1)-quadratic programming.
\newblock {\em Journal of Global Optimization}, 1995.

\bibitem{ragh:88}
P.~Raghavan.
\newblock Probabilistic construction of deterministic algorithms: approximating
  packing integer programs.
\newblock {\em Journal of Computer and System. Sciences}, 1988.

\bibitem{reza:61}
F.~M. Reza.
\newblock {\em An Introduction to Information Theory}.
\newblock Dover Publications, Inc., New York, 1961.

\bibitem{rod:10}
I.~Rodriguez-Lujan, R.~Huerta, Ch. Elkan, and C.~S. Cruz.
\newblock Quadratic programming feature selection.
\newblock {\em Journal of Machine Learning Research}, 2010.

\bibitem{sriva:98}
A.~Srivastav and K.~Wolf.
\newblock Finding dense subgraphs with semidefinite programming.
\newblock In {\em Approximation Algorithms for Combinatiorial Optimization}.
  Springer, 1998.

\bibitem{vafai:93}
H.~Vafaie and K.~De~Jong.
\newblock Robust feature selection algorithms.
\newblock In {\em Proceedings of TAI}, 1993.

\bibitem{yeung:91}
R.W. Yeung.
\newblock A new outlook on {Shannon's} information measures.
\newblock {\em IEEE Trans. on Information Theory}, 1991.

\bibitem{zhao:10}
X.~Zhao, D.~Sun, and K.~Toh.
\newblock A {Newton-CG} augmented {Lagrangian} method for semidefinite
  programming.
\newblock {\em SIAM J. on Optimization}, 2010.

\end{thebibliography}

\
\begin{IEEEbiography}[{\includegraphics[width=25mm,keepaspectratio]{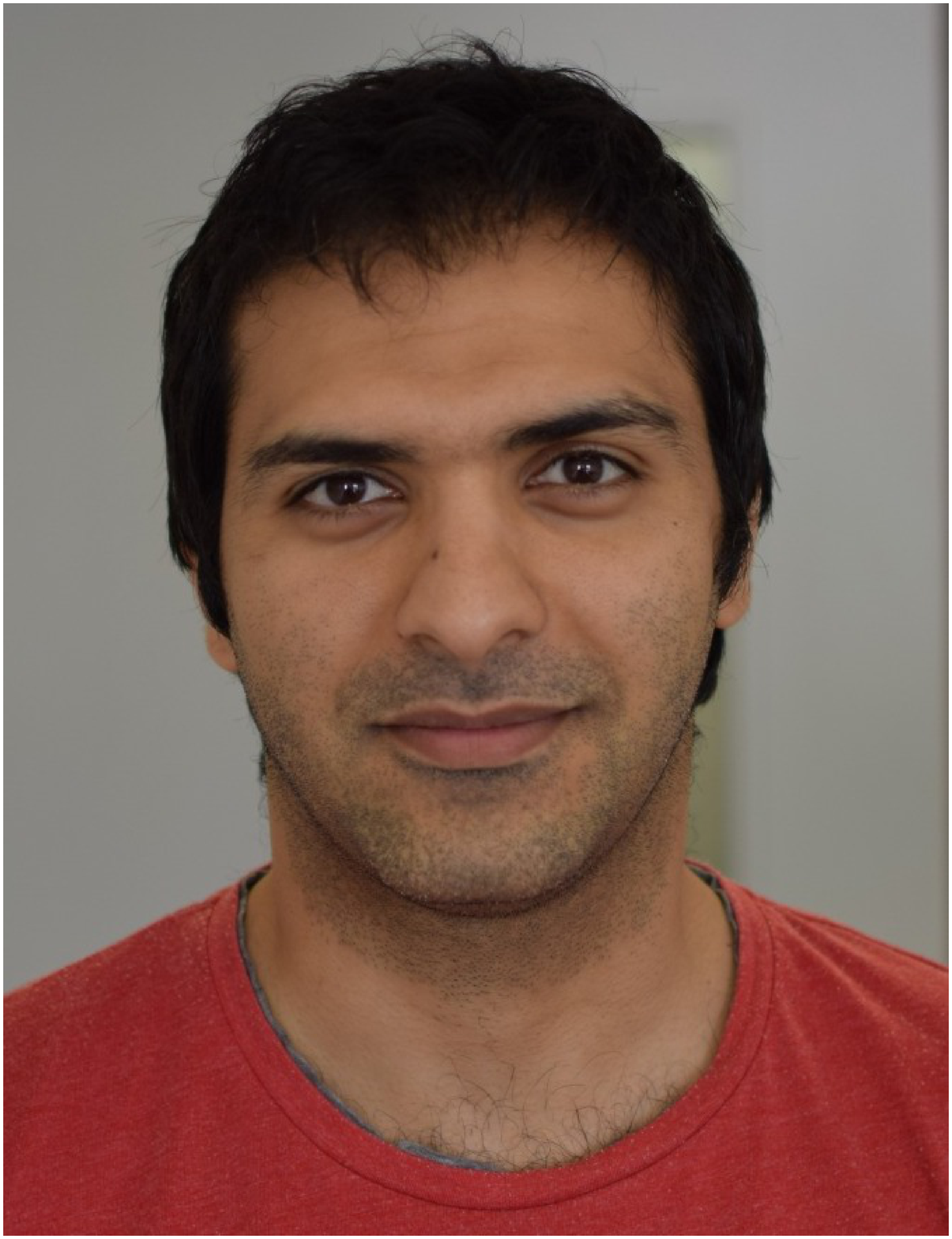}}]%
{Tofigh Naghibi}
received his MSc. degree in electrical engineering from Sharif  University of Technology, Tehran, Iran in 2009. He then joined the speech processing group at ETH Zurich where he is currently working toward a Ph.D. degree in electrical engineering. His research interests include signal processing, pattern recognition and machine learning topics such as boosting methods and online learning.
\end{IEEEbiography}

\begin{IEEEbiography}[{\includegraphics[width=25mm,keepaspectratio]{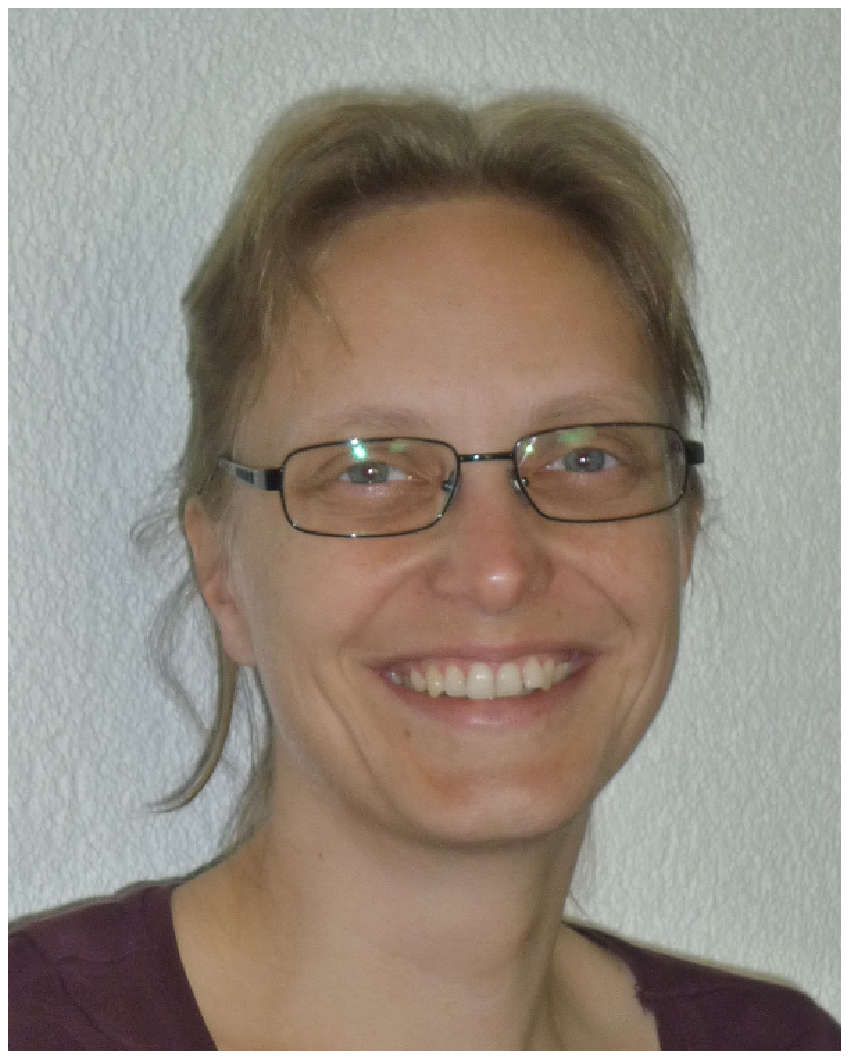}}]%
{Sarah Hoffmann}
received her MSc. in computer science from Dresden technical university in Germany. From 2003 to 2007, she worked for a research group of STMicroelectronics in Rousset, France. She has recently finished her Ph.D. studies in speech processing at ETH Zurich. Her research interests focus on prosody generation in speech synthesis by means of statistical models and semi-supervised speech recognition.
\end{IEEEbiography}
\begin{IEEEbiography}[{\includegraphics[width=25mm,keepaspectratio]{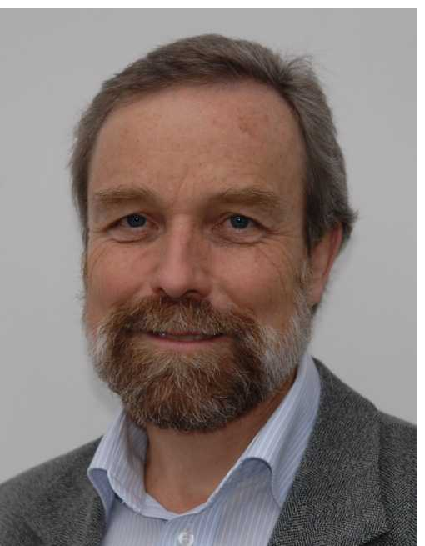}}]%
{Beat Pfister}
 received his diploma in electrical engineering and his Ph.D. from ETH Zurich. Since 1981, he has been the head of the speech processing at the ETH Zurich. He acquired and guided numerous research projects in speech coding, text-to-speech synthesis, speech recognition, and speaker verification. His research interests include text-to-speech synthesis and speech recognition with emphasis on interdisciplinary approaches including signal processing, statistical modelling, knowledge-based systems and linguistics.
\end{IEEEbiography}
\vfill
\end{document}